\documentclass[10pt]{article}
\usepackage[letterpaper, total={6in, 8in}]{geometry}
\usepackage{microtype}
\usepackage{graphicx}
\usepackage{subfigure}
\usepackage{amssymb}
\usepackage{booktabs} 
\usepackage{float}
\usepackage{authblk}

\usepackage{indentfirst}

\usepackage{hyperref}
\usepackage{natbib}

\usepackage[utf8]{inputenc}
\usepackage{hyperref}
\usepackage{graphicx}
\usepackage[utf8]{inputenc}
\usepackage[english]{babel}
\usepackage{xcolor}

\usepackage{amsfonts}
\usepackage{amsmath}

\DeclareMathOperator*{\argmin}{arg\,min}

\usepackage{amsthm}
\newtheorem{theorem}{Theorem}[section]

\newtheorem{Lemma}[theorem]{Lemma}

\theoremstyle{definition}
\newtheorem{remark}[theorem]{Remark}

\newtheorem{innercustomgeneric}{\customgenericname}
\providecommand{\customgenericname}{}
\newcommand{\newcustomtheorem}[2]{
  \newenvironment{#1}[1]
  {
   \renewcommand\customgenericname{#2}
   \renewcommand\theinnercustomgeneric{##1}
   \innercustomgeneric
  }
  {\endinnercustomgeneric}
}

\newcustomtheorem{customthm}{Theorem}
\newcustomtheorem{customlemma}{Lemma}

\title{The Sample Complexity of Meta Sparse Regression}
\author[1]{Zhanyu Wang}
\author[2]{Jean Honorio}
\affil[1,2]{Department of Statistics, Purdue University}
\affil[2]{Department of Computer Science, Purdue University}
\date{}

\begin{document}

\maketitle
\begin{abstract}
This paper addresses the meta-learning problem in sparse linear regression with infinite tasks. We assume that the learner can access several similar tasks. The goal of the learner is to transfer knowledge from the prior tasks to a similar but novel task. For $p$ parameters, size of the support set $k$, and $l$ samples per task, we show that $T \in O((k \log p) / l)$ tasks are sufficient in order to recover the common support of all tasks. With the recovered support, we can greatly reduce the sample complexity for estimating the parameter of the novel task, i.e., $l \in O(1)$ with respect to $T$ and $p$. We also prove that our rates are minimax optimal. A key difference between meta-learning and the classical multi-task learning, is that meta-learning focuses only on the recovery of the parameters of the novel task, while multi-task learning estimates the parameter of all tasks, which requires $l$ to grow with $T$. Instead, our efficient meta-learning estimator allows for $l$ to be constant with respect to $T$ (i.e., few-shot learning).
\end{abstract}

\section{Introduction}

Current machine learning algorithms have shown great flexibility and representational power.
On the downside, in order to obtain good generalization, a large amount of data is required for training.
Unfortunately, in some scenarios, the cost of data collection is high.
Thus, an inevitable question is how to train a model in the presence of few training samples.
This is also called \textbf{Few-Shot Learning} \cite{wang2019few}. Indeed, there might not be much information about an underlying task when only few examples are available. A way to tackle this difficulty is \textbf{Meta-Learning} \cite{vanschoren2018meta}: we gather many similar tasks instead of several examples in one task, and use the data from different tasks to train a model that can generalize well in the similar tasks. This hopefully also guarantees a good performance of the model for a novel task, even when only few examples are available for the new task. In this sense, the model can rapidly adapt to the novel task with prior knowledge extracted from other similar tasks.

As a meta-learning example, for the particular model class of neural networks, 
researchers have developed algorithms such as Matching Networks \cite{vinyals2016matching}, Prototypical Networks \cite{snell2017prototypical}, long-short term memory-based meta-learning \cite{ravi2016optimization}, Model-Agnostic Meta-Learning (MAML) \cite{finn2017model}, among others.
These algorithms are experimental works that have been proved to be successful in some cases.
Unfortunately, there is a lack of theoretical understanding of the generalization of meta-learning, in general, for any model class.
Some of the algorithms can perform very well in tasks related to some specific applications, but it is still unclear why those methods can learn across different tasks with only few examples given for each task. For example, in few shot learning, the case of 5-way 1-shot classification requires the model to learn to classify images from 5 classes with only one example shown for each class. In this case, the model should be able to identify useful features (among a very large learned feature set) in the 5 examples instead of building the features from scratch.

There has been some efforts on building the theoretical foundation of meta-learning. For MAML, \cite{finn2019online} showed that the regret bound in the online learning regime is $O(\log T)$, and \cite{fallah2019convergence} showed that MAML can converge and find an $\epsilon$-first order stationary point with $O(1/\epsilon^2)$ iterations. A natural question is how we can have a theoretical understanding of the meta-learning problem for any algorithm, i.e., the lower bound of the sample complexity of the problem. The upper and lower bounds of sample complexity is commonly analyzed in simple but well-defined statistical learning problems. Since we are learning a novel task with few samples, 
meta-learning falls in the same regime than sparse regression
with large number of covariates $p$ and a small sample size $l$, which is usually solved by $\ell_1$ regularized (sparse) linear regression such as LASSO, albeit for a single task.
Even for a sample efficient method like LASSO, we still need the sample size $l$ to be of order ${\Omega}(k\log p)$ to achieve correct support recovery, where $k$ is the number of non-zero coefficients among the $p$ coefficients. The $l \in {\Theta}(k \log p)$
rate has been proved to be optimal  \cite{wainwright2009sharp}.
If we consider meta-learning, we may be able to bring prior information from similar tasks to reduce the sample complexity of LASSO. In this respect, researchers have considered the multi-task problem, which assumes similarity among different tasks, e.g., tasks share a common support.
Then, one learns for all tasks at once. While it seems that considering many similar tasks together can bring information to each single task, the noise or error is also introduced. {In the results from previous papers, e.g., \cite{jalali2010dirty, obozinski2011support, negahban2011simultaneous}, in order to achieve good performance on all $T$ tasks, one needs the number of samples $l$ to scale with the number of tasks $T$. (See Table \ref{tab:comparison}.) More specifically, one requires $l \in {\Omega}(T)$ or $l \in {\Omega}(\log T)$ for each task, which is not useful in the regime where $l \in O(1)$ with respect to $T$. Results from other papers, e.g., \cite{lounici2009taking, ollier2017regression}, only apply to deterministic (non-random) covariates.}

{
\begin{table}[H]
    \caption{Comparison on rates of $l$ for our meta sparse regression method versus different multi-task learning methods.}\label{tab:comparison}
    \centering
    \begin{tabular}{lll}
    \toprule
        Model & Regularization & Rate of $l$ for support recovery \\
    \midrule
        Ours & ${\ell_1}$ & $O(1)$ (only to recover the common support) \\
        \cite{negahban2011simultaneous} & ${\ell_{1,\infty}}$ & $O(k(T+\log p))$\\
        \cite{obozinski2011support} & ${\ell_{1,2}}$ & $O(kT\log p)$ \\
        \cite{jalali2010dirty} & $\ell_1+\ell_{1,\infty}$ & $O\left(\max(k\log(pT), kT(T+\log p))\right)$ \\
    \bottomrule
    \end{tabular}
\end{table}
}

Our contribution in this paper is as follows. First, we proposed a meta-sparse regression problem and a corresponding generative model that are amenable to solid statistical analysis and also capture the essence of meta-learning.
Second, we prove the upper and lower bounds of the sample complexity of this problem, and show that they match in the sense that $l \in O((k \log p) / T)$ and $l \in \Omega((k \log p) / T)$. Here $p$ is the number of coefficients in one task, $k$ is the number of non-zero coefficients among the $p$ coefficients, and $l$ is the sample size of each task. In short, we assume that we have access to possibly an infinite number of tasks from a distribution of tasks, and for each task we only have limited number of samples. Our goal is to first recover the common support of all tasks and then use it for learning a novel task. { The take-away message of our paper is that} simply by merging all the data from different tasks and solving a $\ell_1$ regularized (sparse) regression problem (LASSO), we can achieve the best sample complexity rate for identifying the common support and learning the novel task. {The merge-and-solve method seems to be intuitive while its validity is not trivial.} To the best of our knowledge, our results are the first to give upper and lower bounds of the sample complexity of meta-learning problems.

\section{Method}
Here, we present the meta sparse regression problem as well as our $\ell_1$ regularized regression method.
\subsection{Problem setting}\label{sec:problem_setting}
We consider the following meta sparse regression model. The dataset containing samples from multiple tasks $\{(X_{t_i,j}, y_{t_i,j}, t_i)|i=1,2,\cdots,T,T+1; j=1,2,\cdots,l\}$ is generated as follows:
\begin{equation}
    y_{t_i,j} = X_{t_i,j}^T (\mathbf{w}^* + \Delta^*_{t_i})  + \epsilon_{t_i,j},
    \label{meta-problem}
\end{equation}
where, $t_i$ indicates the $i$-th task, $\mathbf{w}^* \in \mathbb{R}^p$ is a constant across all tasks, and $\Delta^*_{t_i}$ is the individual parameter for each task. Note that the tasks $\{t_i | i=1,2,\cdots,T\}$ are the related tasks we collect for helping solve the novel task $t_{T+1}$. Each task contains $l$ training samples. The sample size of task $t_{T+1}$ is denoted by $l_{T+1}$, which is equal to $l$ in the setting above, but generally it could also be larger than $l$.

{We assume tasks are related in a way that $\Delta^*_{t_i}$ are independently drawn from one distribution $F^*(\Delta)$, which is a sub-Gaussian distribution with mean 0 and variance proxy $\sigma_\Delta^2$ for each entry. Furthermore, we do not assume entries to be mutually independent.
Sub-Gaussianity is a} very mild assumption, since the class of sub-Gaussian random variables includes for instance Gaussian random variables, any bounded random variable (e.g., Bernoulli, multinomial, uniform), any random variable with strictly log-concave density, and any finite mixture of sub-Gaussian variables.
We denote the support set of each task $t_i (i=1,2,\cdots,T)$ as $S_i=Supp(\mathbf{w}^* + \Delta^*_{t_i})$. 
{Here we consider the case that $S_i\subseteq S=Supp(\mathbf{w}^*)$ and $|S|=k\ll p, k \leq Tl$, $S_{T+1}\subseteq S, |S_{T+1}|=k_{T+1} \leq l$. This assumption is possible as the sub-Gaussian distribution of $\Delta^*_{t_i}$ on the $m$-th entry can be a mixture of some other sub-Gaussian distributions and a Dirac distribution $\delta_{-\mathbf{w}^*_m}$ that can cancel out the $m$-th entry in $\mathbf{w}^*$.}

We assume that $\epsilon_{t_i,j}$ are i.i.d and follow a sub-Gaussian distribution with mean 0 and variance proxy $\sigma_\epsilon^2$.  
{ Sample covariates $X_{t_i,j} \in \mathbb{R}^p$ are independent with each other for any $i, j$. Each sample is a sub-Gaussian vector with variance proxy no greater than $\sigma_x^2$. The samples from different tasks can have different distributions, and we only assume that if we put all samples from all tasks together, their second moment matrix $\Sigma$ satisfies the mutual incoherence condition, i.e., $|||\Sigma_{S^c,S}(\Sigma_{S,S})^{-1}|||_\infty \leq 1-\gamma, \gamma\in(0,1]$ where $\Sigma_{A,B}$ denotes the submatrix of $\Sigma$ with rows in $A$ and columns in $B$. A random vector $X$ in $\mathbb{R}^n$ is sub-Gaussian if for any $z\in\mathbb{R}^n$, the inner product $\langle X,z \rangle$ is a sub-Gaussian random variable. The variance proxy of $X$ is defined as the maximum of the variance proxy of inner product, i.e., $\sigma_x^2 := \sup_{z\in S^{n-1}} \sigma^2_{\langle X,z \rangle}$. This concept $\sigma_x^2$ is proportional to the Orlicz norm of $X$ (introduced in Section \ref{sec:orlicznorm}), which is also used in \cite{vershynin2010introduction}.

\begin{remark}[Difference between meta sparse regression and multitask learning]
Our setting and analysis focuses on the case that the sample size $l$ of each task is fixed and small, and the number of tasks $T$ goes to infinity, while the number of tasks in multitask learning is usually fixed, or grows with the sample size of each task. 
The mutual incoherence condition and $S_i\subseteq S=Supp(\mathbf{w}^*)$ are also common and mild assumptions in the multitask learning literature \cite{jalali2010dirty, negahban2011simultaneous, obozinski2011support}. 
Our problem focuses on recovering only $S$ and $S_{T+1}$ while multitask learning focuses on recovering $S_i$ for all tasks which is much more difficult if the sample size of each task is fixed.
\end{remark}
}

\subsection{Our method}
In meta sparse regression, our goal is to use the prior $T$ tasks and their corresponding data to recover the common support of all tasks. We then estimate the parameters for the novel task. For the setting we explained above, this is equivalent to recover $(\mathbf{w}^*, \Delta^*_{t_{T+1}})$.

First, we determine the common support $S$ over the prior tasks $\{t_i | i=1,2,\cdots,T\}$ by the support of $\hat{w}$ formally introduced below, i.e., $\hat{S}=Supp(\hat{w})$, where
\begin{equation}
    \ell(\mathbf{w}) = \frac{1}{2Tl}\sum_{i=1}^T \sum_{j=1}^{l} \|y_{t_i,j} -  X_{t_i,j}^T \mathbf{w}\|_2^2,\quad \hat{\mathbf{w}} = \argmin_{\mathbf{w}} \left\{\ell(\mathbf{w}) + \lambda\|\mathbf{w}\|_1 \right\}
    \label{est_w}
\end{equation}
Note that we have $T$ tasks in total, and $l$ samples for each task.

Second, we use the support $\hat{S}$ as a constraint for recovering the parameters of the novel task $t_{T+1}$. That is
\begin{equation}
    \ell_{T+1}(\mathbf{w}) = \frac{1}{2l} \sum_{j=1}^{l} \|y_{t_{T+1},j} -  X_{t_{T+1},j}^T \mathbf{w}\|_2^2,\  \hat{\mathbf{w}}_{T+1} = \argmin_{\mathbf{w}, Supp(\mathbf{w})\subseteq\hat{S}} \left\{\ell_{T+1}(\mathbf{w}) + \lambda_{T+1}\|\mathbf{w} \|_1 \right\}
    \label{est_new_w}
\end{equation}

We point out that our method makes a proper application of $\ell_1$ regularized (sparse) regression, and in that sense is somewhat intuitive. In what follows, we show that this method correctly recovers the common support and the parameter of the novel task.
At the same time, our method is minimax optimal, i.e., it achieves the optimal sample complexity rate.

\section{Main results}

First, we state our result for the recovery of the common support among the prior $T$ tasks.

\begin{theorem}\label{theorem_meta}
{
Let $\hat{\mathbf{w}}$ be the solution of the optimization problem (\ref{est_w}). If $T \in \Omega\left(\frac{k\log(p-k)}{l}\right)$ and  
$$
\lambda\in \Omega \left(\max\left(\sigma_\epsilon \sigma_x\sqrt{\frac{\log(p-k)}{Tl}}, \sigma_\Delta \sigma_x^2\sqrt{\frac{\log(p-k)k^5l^3}{T}}, \sigma_\Delta \sigma_x^2\sqrt{\frac{(\log(p-k))^3 k^5l^3}{T^2}}\right)\right)
$$
with probability greater than 
$1-c_1 \mathrm{exp}(-c_2 \min(\log(p-k)))$}, we have that 
\begin{enumerate}
    \item the support of $\hat{\mathbf{w}}$ is contained within $S$ (i.e., $S(\hat{\mathbf{w}}) \subseteq S$);
    \item $\|\hat{\mathbf{w}} - \mathbf{w}^*\|_\infty \leq { c_3 \lambda \sqrt{k}}$, 
\end{enumerate} 
where $c_1,c_2,c_3$ are constants.
\end{theorem}

{
\begin{remark}[Theorem \ref{theorem_meta}]
The scale terms $k \log(p-k)$ in $T$ and $\sqrt{\log(p-k)/Tl}$ in $\lambda$ are typically encountered in the analysis of the single-task sparse regression or LASSO \cite{wainwright2009sharp}. The two additional terms in $\lambda$ are due to the difference in the coefficients among tasks, and our simulations show that $\lambda=\sqrt{\log(p-k)/Tl}$ is sufficient in the settings we consider. In our technical analysis, the additional terms come from the concentration inequality of a random variable with finite Orlicz norm we used, which is the main novelty in our proof: bounding the product of three random variables.
\end{remark}
}

Next, we state our result for the recovery of the parameters of the novel task. The proof can be found in appendix Section \ref{app:sec_proof_novel}.

\begin{theorem}\label{theorem_new}
Let $\hat{\mathbf{w}}_{T+1}$ be the solution of the optimization problem (\ref{est_new_w}). With the support $\hat{S}$ recovered from Theorem \ref{theorem_meta}, if $k'=k_{T+1}$, $\lambda' = \lambda_{T+1} \in {\Theta\left(\sigma_\epsilon \sigma_x\sqrt{{\log(k-k')}/{l}}\right)}$ and 
$l \in {\Omega}\left({k'\log(k-k')}\right)$, with probability greater than {$1-c'_1 \mathrm{exp}(-c'_2 \log(k-k'))$},  we have that
\begin{enumerate}
    \item the support of $\hat{\mathbf{w}}_{T+1}$ is contained within $S_{T+1}$ (i.e., $S(\hat{\mathbf{w}}_{T+1}) \subseteq S_{T+1} \subseteq S$);
    \item $\|\hat{\mathbf{w}}_{T+1} - (\mathbf{w}^* + \Delta^*_{t_{T+1}})\|_\infty \leq  { c_3' \lambda'\sqrt{k'} }$, 
\end{enumerate} 
where $c'_1, c'_2, c'_{3}$ are constants.
\end{theorem}

The theorems above provide an upper bound of the sample complexity, which can be achieved by our method. 
The lower bound of the sample complexity is an information-theoretic result, and it relies on the construction of a restricted class of parameter vectors.
We consider a special case of the setting we previously presented: all non-zero entries in $\mathbf{w}^*$ are 1, 
and all non-zero entries in $\mathbf{w}^* + \Delta^*_{t_i}$ are also 1. 
We use $\Theta$ to denote the set of all possible parameters $\theta^* = (\mathbf{w}^*, \Delta_{t_{T+1}}^*)$. 
Therefore the number of possible outcomes of the parameters  $|\Theta|={p\choose k}{k\choose k_{T+1}} \in O(p^k k^{k_{T+1}})$. 

If the parameter $\theta^*$ is chosen uniformly at random from $\Theta$, for any algorithm estimating this parameter by $\hat\theta$, the answer is wrong (i.e., $\hat\theta \neq \theta^*$) with probability greater than $1/2$ if $O(T l + l_{T+1}) \lesssim k\log p + k_{T+1}\log k$. Here we use $l_{T+1}$ to denote the sample size of task $t_{T+1}$. This fact is proved in the following theorem. The detailed proof is included in appendix Section \ref{app:sec_proof_fano}.

\begin{theorem}\label{theorem_fano}
Let $\Theta := \{\theta = (\mathbf{w}, \Delta_{t_{T+1}}) | \mathbf{w} \in \{0,1\}^p,  \|\mathbf{w}\|_0=k, \Delta_{t_i} \in \{0,-1\}^p, Supp(\Delta_{t_i}) \subseteq Supp(\mathbf{w}),  \|\mathbf{w} + \Delta_{t_i}\|_0 = k_i \}$. Furthermore, assume that $\theta^* = (\mathbf{w}^*, \Delta^*_{t_{T+1}})$ is chosen uniformly at random from $\Theta$. We have:
$$
\mathbb{P}[\hat\theta \neq \theta^*] \geq 1 - \frac{\log 2 + c''_1\cdot Tl + c''_2\cdot l_{T+1}}{\log |\Theta|}
$$
where $c''_1,c''_2$ are constants.
\end{theorem}

In the following section, we prove that the mutual information $\mathbb{I}(\theta^*, S)$ between the true parameter $\theta^*$ and the data $S$ is bounded by $c''_1\cdot Tl + c''_2\cdot l_{T+1}$. 
In order to prove Theorem 3.3, we use Fano's inequality and the construction of a restricted class of parameter vectors.
The use of Fano's inequality and restricted ensembles is customary for information-theoretic lower bounds \cite{wang2010information, santhanam2012information, tandon2014information}.

Note that from Theorem \ref{theorem_fano}, we know if $T{\in o}(\frac{k\log p}{l})$ and $l_{T+1}{\in o}(k_{T+1}\log k)$, then any algorithm will fail to recover the true parameter very likely. On the other hand, if we have $T \in {\Omega}(\frac{k\log p}{l})$ and $l_{T+1} \in {\Omega}(k_{T+1}\log k)$, by Theorem \ref{theorem_meta} and \ref{theorem_new}, we can recover the support of $\mathbf{w}^*$ and $\Delta^*_{T+1}$ (by $\mathbf{w}^*_{T+1} - \mathbf{w}^*$). Therefore we claim that our rates of sample complexity is minimax optimal.

\section{Sketch of the proof of Theorem \ref{theorem_meta}}

We use the primal-dual witness framework \cite{wainwright2009sharp} to prove our results. 
First we construct the primal-dual candidate; then we show that the construction succeeds with high probability.
Here we outline the steps in the proof. (See the supplementary materials for detailed proofs.)

We first introduce some useful notations:

$\mathbf{X}_{t_i} \in \mathbb{R}^{l\times p}$ is the matrix of collocated $X_{t_i, j}$ (covariates of all samples in the $i$-th task). Similarly, $\mathbf{y}_{t_i} \in \mathbb{R}^{l}$ and $\mathbf{\epsilon}_{t_i} \in \mathbb{R}^{l}$. 
$\mathbf{X}_{[T]} \in \mathbb{R}^{Tl\times p}$ is the matrix of collocated $\mathbf{X}_{t_i}$ (covariates of all samples in all tasks). Similarly, $\mathbf{\epsilon}_{[T]} \in \mathbb{R}^{Tl}$.
$\mathbf{X}_{t_i, S} \in \mathbb{R}^{l\times k}$ is the sub-matrix of $\mathbf{X}_{t_i}$ containing only the rows corresponding to the support of $\mathbf{w}^*$, i.e., $S$ with $|S|=k$. Similarly, $\mathbf{X}_{[T], S} \in \mathbb{R}^{Tl\times k}$, $\Delta_{t_i, S}^* \in \mathbb{R}^{k}$, and $\mathbf{w}_S \in \mathbb{R}^{k}$.
$\mathbf{A}_{S, S} \in \mathbb{R}^{k\times k}$ is the sub-matrix of $\mathbf{A} \in \mathbb{R}^{p\times p}$ containing only the rows and columns corresponding to the support of $\mathbf{w}^*$.

\subsection{Primal-dual witness}
\indent\textbf{Step 1:} Prove that the objective function has positive definite Hessian when restricted to the support, i.e., $\forall \mathbf{w}_{S^c} = \mathbf{0}, \forall\mathbf{w}_S \in \mathbb{R}^{|S|},\ [\nabla^2 \ell((\mathbf{w}_S,\mathbf{0}))]_{S,S} \succ 0$

\textbf{Step 2:} 
Set up a restricted problem:
\begin{equation}
    \tilde{\mathbf{w}}_S = \argmin_{\mathbf{w}_S\in\mathbb{R}^{|S|}} \ell((\mathbf{w}_S, \mathbf{0})) + \lambda\|\mathbf{w}_S\|_1
    \label{restricted_problem}
\end{equation}

\textbf{Step 3:} 
Choose the corresponding dual variable $\tilde{\mathbf{z}}_S$ to fulfill the complementary slackness condition:

$\forall i \in S$, $\tilde{\mathbf{z}}_i=sign(\tilde{\mathbf{w}}_i)$ if $\tilde{\mathbf{w}}_i \neq 0$, otherwise $\tilde{\mathbf{\mathbf{z}}}_i\in[-1,+1]$

\textbf{Step 4:} 
Solve $\tilde{\mathbf{z}}_{S^c}$ to let $(\tilde{\mathbf{w}}, \tilde{\mathbf{z}})$ fulfill the stationarity condition:
\begin{align}
        [\nabla \ell((\tilde{\mathbf{w}}_S,\mathbf{0}))]_S &+ \lambda\tilde{\mathbf{z}}_S = 0 \label{stationarity} \\ 
    [\nabla \ell((\tilde{\mathbf{w}}_S,\mathbf{0}))]_{S^c} &+ \lambda\tilde{\mathbf{z}}_{S^c} = 0 \label{stationarity2}
\end{align}

\textbf{Step 5:} 
Verify that the strict dual feasibility condition is fulfilled for $\tilde{\mathbf{z}}_{S^c}$:
$$
\|\tilde{\mathbf{z}_{S^c}}\|_\infty < 1
$$

{In order to prove support recovery, we only need to show that \textbf{step 1} and \textbf{step 5} hold. The proof of \textbf{step 1} being satisfied with high probability under the condition $T \in O(k/l)$ is in appendix Section \ref{app:step1}. Next we show that \textbf{step 5} also holds with high probability.}

\subsection{Strict dual feasibility condition}\label{sec:z_sc}
We first rewrite (\ref{stationarity}) as follows:
\begin{align*}
\frac{1}{Tl}\sum_{i=1}^T \mathbf{X}_{t_i, S}^T\mathbf{X}_{t_i, S} (\tilde{\mathbf{w}}_S - {\mathbf{w}}_S^*) = - \lambda\tilde{\mathbf{z}}_S + \frac{1}{Tl}\sum_{i=1}^T \mathbf{X}_{t_i, S}^T\epsilon_{t_i} + \frac{1}{Tl}\sum_{i=1}^T \mathbf{X}_{t_i, S}^T\mathbf{X}_{t_i, S} \Delta_{t_i, S}^*
\end{align*}

Then we solve for $(\tilde{\mathbf{w}}_S - {\mathbf{w}}_S^*)$. 
and plug it in (\ref{stationarity2}). We have

\begin{align*}
    \tilde{\mathbf{z}}_{S^c} = & \underbrace{\mathbf{X}_{[T], S^c}^T \left\{\frac{1}{Tl}\mathbf{X}_{[T], S}(\hat\Sigma_{S,S})^{-1}\tilde{\mathbf{z}}_S + \Pi_{\mathbf{X}_{[T],S}^{\perp}} \left(\frac{\epsilon_{[T]}}{\lambda Tl}\right) \right\}}_{\tilde{\mathbf{z}}_{S^c, 1}} + \underbrace{\frac{1}{\lambda Tl}\sum_{i=1}^T X_{t_i, S^c}^T X_{t_i, S} \Delta_{t_i,S}^*}_{\tilde{\mathbf{z}}_{S^c, 2}} \\
        & - \underbrace{\frac{1}{\lambda (Tl)^2} \mathbf{X}_{[T], S^c}^T \mathbf{X}_{[T], S}(\hat\Sigma_{S,S})^{-1} \left( \sum_{i=1}^T X_{t_i, S}^T X_{t_i, S} \Delta_{t_i,S}^* \right)}_{\tilde{\mathbf{z}}_{S^c, 3}} \\
\end{align*}
where $\Pi_{\mathbf{X}_{[T],S}^{\perp}} := I_{n\times n} - \mathbf{X}_{[T],S}(\mathbf{X}_{[T],S}^T \mathbf{X}_{[T],S})^{-1}\mathbf{X}_{[T],S}^T$ is an orthogonal projection matrix, { $\hat\Sigma_{S,S}=\frac{1}{Tl}\sum_{i=1}^T \mathbf{X}_{t_i, S}^T\mathbf{X}_{t_i, S}$ is the sample covariance matrix, and $\tilde{\mathbf{z}}_S$ is the dual variable chosen at step 3.}

{
One can bound the $\ell_\infty$ norm of $\tilde{\mathbf{z}}_{S^c, 1}$ by the techniques used in \cite{wainwright2009sharp}: if $\lambda\in {\Omega}\left(\sigma_\epsilon \sigma_x \sqrt{\frac{\log(p-k)}{Tl}}\right)$ and $T \in \Omega\left(\frac{k\log(p-k)}{l}\right)$, we have
$$
\mathbb{P}[\|\tilde{\mathbf{z}}_{S^c, 1}\|_\infty \geq 1 - \gamma/2] \leq 2e^{-c_4 \log(p-k)}.
$$
where $c_4$ is a constant. Proof of this result is shown in appendix Section \ref{app:bound1}. 
}

Note that the remaining two terms $\tilde{\mathbf{z}}_{S^c, 2}, \tilde{\mathbf{z}}_{S^c, 3}$ containing $\Delta_{t_i}$ are new to the meta-learning problem and need to be handled with novel proof techniques. 

We first rewrite $\tilde{\mathbf{z}}_{S^c, 2}$ with respect to each of its entries (denoted by $\tilde{\mathbf{z}}_{n, 2}$) as follows: $\forall n \in S^c$, we have
\begin{equation}
    \tilde{\mathbf{z}}_{n, 2} = \frac{1}{\lambda Tl}\sum_{i=1}^T\sum_{j=1}^l\sum_{q\in S} X_{t_i, j, n} X_{t_i, j, q} \Delta_{t_i,q}^*
    \label{zn2}
\end{equation}

We know that $X_{t_i, j, n}, X_{t_i, j, q}, \Delta_{t_i,q}^*$ are sub-Gaussian random variables. It is well-known that the product of two sub-Gaussian is sub-exponential (whether they are independent or not). To characterize the product of three sub-Gaussians and the sum of the i.i.d. products, we need to use Orlicz norms and a corresponding concentration inequality.

\subsection{Orlicz norm}\label{sec:orlicznorm}

Here we introduce the concept of exponential Orlicz norm. For any random variable $X$ and $\alpha > 0$, we define the $\psi_\alpha$ (quasi-) norm as

$$
\|X\|_{\psi_\alpha} = \inf \left\{ t>0: \mathbb{E}\exp \left({|X|^\alpha}/{t^\alpha}\right)\leq 2\right\}
$$
We define $\inf \emptyset = \infty$. This concept is a generalization of sub-Gaussianity and sub-exponentiality since the random variable family with finite exponential Orlicz norm $\|\cdot\|_{\psi_\alpha}$ corresponds to the $\alpha$-sub-exponential tail decay family which is defined by

$$
\mathbb{P}\left(|X|\geq t\right) \leq c\exp \left(-{t^\alpha}/{C}\right)\ \forall t \geq 0.
$$
where $c, C$ are constants. More specifically, if $\|X\|_{\psi_\alpha} = k$, we set $c=2, C=k^\alpha$ so that $X$ fulfills the $\alpha$-sub-exponential tail decay property above. We have two special cases of Orlicz norms: $\alpha=2$ corresponds to the family of sub-Gaussian distributions {(with $\|X\|_{\psi_2} = \Theta(\sigma_x)$, i.e., proportional to the variance proxy)} and $\alpha=1$ corresponds to the family of sub-exponential distributions. 

A good property of the Orlicz norm is that the product or the sum of many random variables with finite Orlicz norm has finite Orlicz norm as well (possibly with a different $\alpha$.) We state this property in the two lemmas below.

\begin{Lemma}\label{orlicz_prod_sum}[Lemma A.1, A.3 in \cite{gotze2019concentration}]
Let $X_1, \cdots, X_k$ be random variables such that $\|X_i\|_{\psi_{\alpha_i}} < \infty$ for some $\alpha_i\in(0,1]$ and let $t=\frac{1}{\sum_{i=1}^k \alpha_i^{-1}}$. Then 
$$
\left\|\prod_{i=1}^k X_i\right\|_{\psi_t} \leq \prod_{i=1}^k \|X_i\|_{\psi_{\alpha_i}}.\quad \text{Moreover, } \left\|\sum_{i=1}^k X_i\right\|_{\Psi_\alpha} \leq k^{1/\alpha}\left(\sum_{i=1}^k\|X_i\|_{\Psi_\alpha}\right) \text{ if }\  \forall i, \alpha_i = \alpha.
$$
\end{Lemma}

By the lemma above, we know that the sum (with respect to $j,m$) of the products in (\ref{zn2}) is a $\frac{2}{3}$-sub-exponential tail decay random variable. The details are shown in the next subsection. {This result does not require any independence condition, thus we will use this fact for bounding both $\tilde{\mathbf{z}}_{S^c,2}$ and $\tilde{\mathbf{z}}_{S^c,3}$.}

\subsection{$\frac{2}{3}$-sub-exponential tail decay random variable}

For $j\in S^c, q\in S$, we have

{
$$
\tilde{\mathbf{z}}_{j, 2} = \frac{1}{\lambda Tl}\sum_{i=1}^T \left( \sum_{m=1}^l  \sum_{q\in S}\Delta_{t_i,q}^* X_{t_i, j, m} X_{t_i, q, m} \right) := \frac{1}{T} \sum_{i=1}^T  \tilde{\mathbf{z}}_{j, 2, i}
$$

From Lemma \ref{orlicz_prod_sum}, we know 
$$
\begin{aligned}
    \left\|\Delta_{t_i,q}^* X_{t_i, j, m} X_{t_i, q, m} \right\|_{\psi_{\frac{2}{3}}} &\leq \|\Delta_{t_i,q}^*\|_{\psi_{2}}  \|X_{t_i, j, m}\|_{\psi_{2}} \|X_{t_i, q, m}\|_{\psi_{2}} = c_5 \sigma_\Delta \sigma_x^2 \\
    \left\|\frac{1}{\lambda l}\sum_{m=1}^l \sum_{q\in S}\Delta_{t_i,q}^* X_{t_i, j, m} X_{t_i, q, m}\right\|_{\psi_{\frac{2}{3}}} &\leq \frac{(kl)^{\frac{5}{2}} c_5 \sigma_\Delta \sigma_x^2}{\lambda l} = \frac{c_5 \sigma_\Delta \sigma_x^2 k^{\frac{5}{2}}l^{\frac{3}{2}}}{\lambda}.
\end{aligned}
$$
where $c_5$ is a constant.

\subsection{Concentration inequality for $\tilde{\mathbf{z}}_{S^c, 2}$}

We know that for different $q\in S$ and $i$, the random variables $\tilde{\mathbf{z}}_{j, 2, i}$ are independent with $\mathbb{E}\tilde{\mathbf{z}}_{j, 2, i} = 0$ and $\|\tilde{\mathbf{z}}_{j, 2, i}\|_{\psi_{\alpha_{2/3}}} \leq c_5\sigma_\Delta \sigma_x^2 k^{\frac{5}{2}}l^{\frac{3}{2}}/\lambda$. Now we use a concentration inequality to bound $\tilde{\mathbf{z}}_{j, 2}$.

\begin{Lemma}[Theorem 1.4 in \cite{gotze2019concentration}]\label{lem:concentration}
Let $X_1, \cdots, X_k$ be a set of independent random variables satisfying $\|X_i\|_{\psi_{\alpha_{2/3}}} \leq M$ for some $M > 0$. There exists a constant $C_3$ such that for any $t>0$, we have
{
$$
\mathbb{P}\left(\left|\frac{1}{n}\sum_{i=1}^n (X_i-\mathbb{E}X_i)\right| \geq t\right) \leq 2\mathrm{exp}\left(-\frac{f_3(M,t,n)}{C_3}\right)
$$}
where
$$
f_3(M,t,n) = \min\left(\frac{t^2n}{M^2}, \frac{tn}{M}, \left(\frac{tn}{M}\right)^{\frac{2}{3}} \right).
$$
\end{Lemma}

We let $t = \frac{\gamma}{4}, n = T, T\in\Omega\left(\frac{k\log (p-k)}{l}\right)$, and 
$$\lambda\in \Omega \left(\sigma_\Delta \sigma_x^2\cdot \max\left(\sqrt{\frac{\log(p-k)k^5l^3}{T}}, \sqrt{\frac{(\log(p-k))^3 k^5l^3}{T^2}}\right)\right).
$$
Then we have 
$$
\begin{aligned}
\mathbb{P}\left(\left|\tilde{\mathbf{z}}_{j, 2} \right| \geq \frac{\gamma}{4}\right) &= \mathbb{P}\left(\left|\frac{1}{T}\sum_{i=1}^T  \tilde{\mathbf{z}}_{j, 2, i}\right| \geq \frac{\gamma}{4}\right) = O(\mathrm{exp}(-\log (p-k)))
\end{aligned}
$$

Therefore, $\|\tilde{\mathbf{z}}_{S^c, 2}\|_\infty$ can be bounded by $\gamma/4$ with probability
$$
\mathbb{P}[\|\tilde{\mathbf{z}}_{S^c, 2}\|_\infty \geq \gamma/4] \leq c_6\mathrm{exp}(-c_7\log(p-k))
$$
where $c_6,c_7$ are constants.
}

\subsection{Bound on  $\|\tilde{\mathbf{z}}_{S^c, 3}\|_\infty$}\label{z3}

By definition,
$$
\begin{aligned}
&\tilde{\mathbf{z}}_{S^c, 3} = \frac{1}{\lambda (Tl)^2} \mathbf{X}_{[T], S^c} \mathbf{X}_{[T], S}(\hat\Sigma_{S,S})^{-1} \left( \sum_{i=1}^T X_{t_i, S}^T X_{t_i, S} \Delta_{t_i,S}^* \right) \\
& = \frac{1}{Tl} \mathbf{X}_{[T], S^c} \mathbf{X}_{[T], S}(\hat\Sigma_{S,S})^{-1} \frac{1}{\lambda Tl}\left( \sum_{i=1}^T X_{t_i, S}^T X_{t_i, S} \Delta_{t_i,S}^* \right) := \frac{1}{Tl} \mathbf{X}_{[T], S^c} \mathbf{X}_{[T], S}(\hat\Sigma_{S,S})^{-1} \zeta_S
\end{aligned}
$$
where we define
\begin{equation}\label{eq:zeta_S}
\zeta_S := \frac{1}{\lambda Tl}\left( \sum_{i=1}^T X_{t_i, S}^T X_{t_i, S} \Delta_{t_i,S}^* \right).
\end{equation}

{
Since the independence between random variables is not necessary in Lemma \ref{orlicz_prod_sum}, we use the same technique for bounding $\tilde{\mathbf{z}}_{S^c, 2}$ to bound $\zeta_S$. More specifically, under the same condition to bound $\|\tilde{\mathbf{z}}_{S^c, 2}\|_\infty$, we have
\begin{equation}\label{eq:bound_zeta_S}
\mathbb{P}\left[\|\zeta_S\|_\infty \geq \frac{\gamma}{2-\gamma}\right] \leq  c_8\mathrm{exp}(-c_9\log k)
\end{equation}

Now we can transform $\tilde{\mathbf{z}}_{S^c, 3}$ into the first part in $\tilde{\mathbf{z}}_{S^c, 1}$ by replacing $\zeta_S$ with $\tilde{\mathbf{z}}_S$. We know that $\|\tilde{\mathbf{z}}_S\|_\infty \leq 1$. Therefore we can bound $\tilde{\mathbf{z}}_{S^c, 3}$ using the same technique for bounding $\|\tilde{\mathbf{z}}_{S^c, 1}\|_\infty$ in appendix Section \ref{app:bound1}: if $T \in \Omega\left( \frac{k\log(p-k)}{l}\right)$, we have 
$$
\mathbb{P}[\|\tilde{\mathbf{z}}_{S^c, 3}\|_\infty \geq \gamma/2] \leq c_{10}\mathrm{exp}(-c_{11}\log(p-k)).
$$
where $c_8,c_9,c_{10},c_{11}$ are constants.
}

{
\subsection{Bound on  $\|\tilde{\mathbf{z}}_{S^c}\|_\infty$ and the estimation error $\|\hat{\mathbf{w}} - \mathbf{w}^*\|_\infty$}

Since we have bounded each part of $\tilde{\mathbf{z}}_{S^c}$, we have $\|\tilde{\mathbf{z}}_{S^c}\|_\infty < 1$ with high probability, therefore the first part of Theorem \ref{theorem_meta} about support recovery ($S(\hat{\mathbf{w}}) \subseteq S$) is proved through primal-dual witness by finishing \textbf{step 1} and \textbf{step 5}. The proof for the second part of Theorem \ref{theorem_meta} about the estimation error uses similar techniques. Details can be found in the appendix Section \ref{app:bound_err}.
}

\section{Discussions}

\begin{figure}[H]
  \begin{minipage}[c]{0.6\textwidth}
    \includegraphics[width=\textwidth]{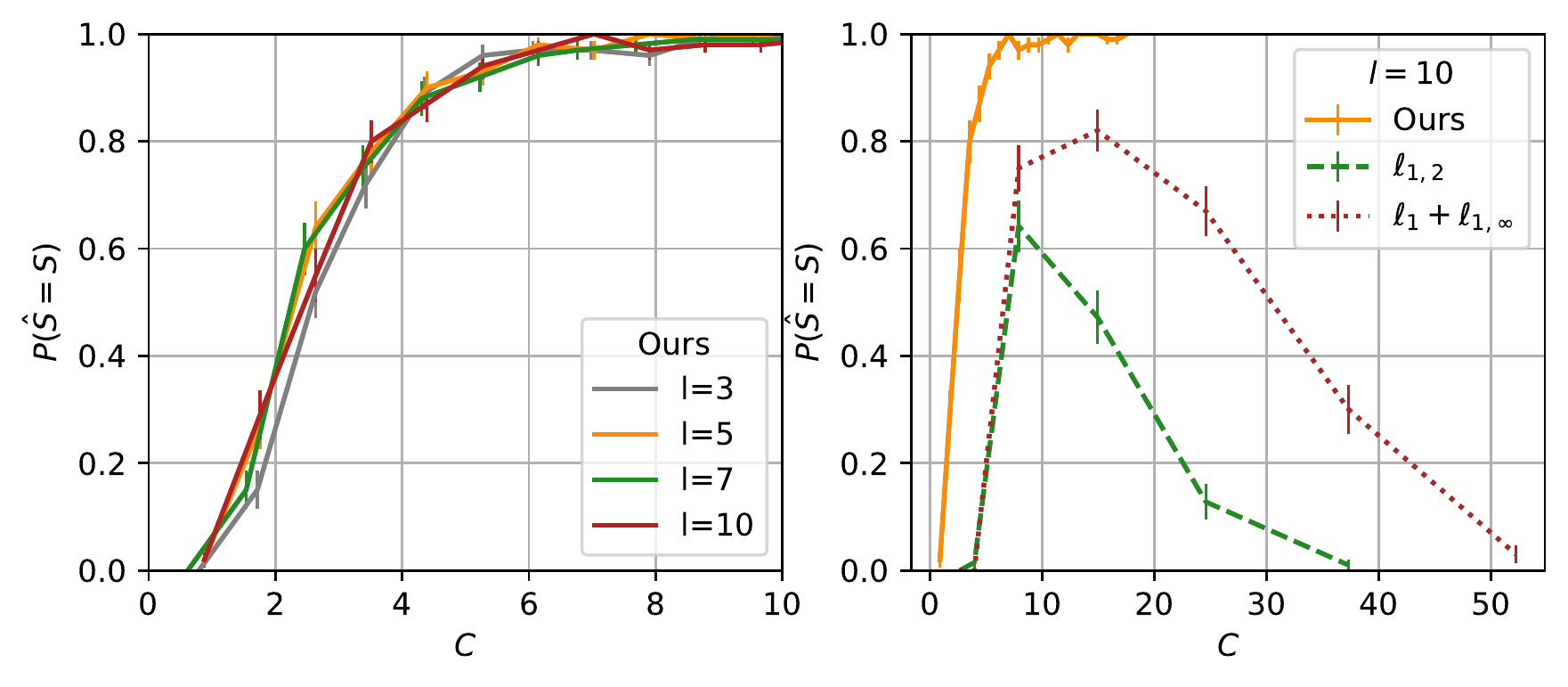}
  \end{minipage}\hfill
  \begin{minipage}[c]{0.39\textwidth}
    \caption{
       Simulations for Theorem \ref{theorem_meta} on the probability of exact support recovery. We use $\lambda=\sqrt{\frac{\log p }{Tl}}$.
       \textbf{Left:} Probability of exact support recovery for different number of tasks under various settings of $l$. The x-axis is set by $C:=\frac{Tl}{k\log(p-k)}$.  \textbf{Right:} Our method outperforms multi-task methods. 
    } \label{fig:simulation}
  \end{minipage}
\end{figure}

Our problem setting and method are amenable to solid statistical analysis. By focusing on sparse regression, our analysis shows clearly the difference between meta-learning and multi-task learning. In meta-learning, we only need to recover $\mathbf{w}^*$ and $\Delta^*_{t_{T+1}}$, thus the number of samples needed for each task (including the novel task) is $l\in O((k\log p)/T + k_{T+1}\log k)$. When $T\rightarrow \infty$, {meta-learning can recover $\mathbf{w}^*$ with high probability (shown in the left panel of Figure \ref{fig:simulation} where $C:=\frac{Tl}{k\log(p-k)}$), therefore for the novel task, it} only needs $l\in O(k_{T+1}\log k)$. For multi-task learning, one needs to recover $(\mathbf{w}^* + \Delta^*_{t_i})$ for all $t_i$, which requires the sample size at least $l \in O(k(T + \log p))$
(see Table \ref{tab:comparison}.) When $T\rightarrow \infty$, the sample size of multi-task learning goes to infinity { which is supported by the right panel of Figure \ref{fig:simulation}: when $T$ grows and $l$ is fixed at $10$, the probability of exact support recovery of $S=\bigcup_{i=1}^T S_i$ decreases to $0$.} (For details about Figure \ref{fig:simulation} and additional experiments, see appendix Section \ref{app:experiment}.)

While meta sparse regression might apparently look similar to the classical sparse random effect model \cite{bondell2010joint}, a key difference is that in the random effect model, the experimenter is interested on the distribution of the estimator $\mathbf{w}^*$ instead of support recovery. 
To the best of our knowledge, our results are the first to give upper and lower bounds of the sample complexity of meta-learning problems.

Although our paper shows that a proper application of $\ell_1$ regularized (sparse) regression achieves the minimax optimal rate, it is still unclear whether there is a method that can improve the constants in our results. To have further theoretical understanding of meta-learning, 
one could consider other algorithms, such as nonparametric regression or neural networks. We believe that our results are a solid starting point for the sound statistical analysis of meta-learning.

\bibliography{main}
\bibliographystyle{apalike}

\onecolumn
\appendix

\section{Proof of step 1 in the primal-dual witness}\label{app:step1}
We know that 
\begin{equation}
    [\nabla^2 \ell((\mathbf{w}_S,\mathbf{0}))]_{S,S} \succ 0 \Leftrightarrow \frac{1}{Tl}[\mathbf{X}_{[T]}^T \mathbf{X}_{[T]}]_{S,S} \succ 0
    \label{step1_x}
\end{equation}
{

We first show a useful theorem on bounding the difference between the sample covariance matrix and the population covariance matrix.
\begin{Lemma}[Theorem 4.6.1 in \cite{vershynin2018high}]\label{vershynin461}
Let $A$ be an $m\times n$ matrix whose rows $A_i$ are independent, mean zero, sub-gaussian isotropic random vectors in $\mathbb{R}^n$. Then for any $t\geq 0$, we have
$$
\left\|\frac{1}{m} A^T A - I_n\right\| \leq K^2 \max(\delta,\delta^2) \quad \text{ where } \delta = C\left(\sqrt{\frac{n}{m}} + \frac{t}{\sqrt{m}}\right), K = \max_i \|A_i\|_{\psi_2}
$$
with probability $1-2e^{-t^2}$.
\end{Lemma}

To prove (\ref{step1_x}), we need to bound $\lambda_{min}([\frac{1}{Tl}[\mathbf{X}_{[T]}^T \mathbf{X}_{[T]}]_{S,S})$ away from 0. We find independent isotropic random vectors $Z_i$ such that $X_i = \Sigma_{S,S}^{1/2} Z_i$ (by the proof of Theorem 4.7.1 in \cite{vershynin2018high}, $\Sigma_{S,S}$ is the submatrix of $\Sigma$ restricted on $S$ in both row and column, and $Z_i$ are also sub-Gaussian with $\|Z_i\|_{\psi_2}\leq K$) and define $A = [Z_1, Z_2, \cdots, Z_m]^T$, $m=Tl, n=k$. Then we use a similar technique as Lemma 4.1.5 in  \cite{vershynin2018high}: 
$$
K^2\max (\delta, \delta^2) \geq \left\|\frac{1}{m}A^T A - I_n\right\| \geq \left|\left\langle \left(\frac{1}{m}A^TA - I_n\right)x, x \right\rangle\right| = \left|\lambda_{min}\left(\frac{1}{m}A^TA\right) - 1\right|
$$
where we set $x = \argmin_{a\in \mathbb{S}^{n-1}} a^T (A^TA) a$. 

If $K^2\max (\delta, \delta^2) < 1/2$, we will have $\lambda_{min}(\frac{1}{m}A^TA) > 1/2$ and also $\lambda_{min}(\frac{1}{Tl}[\mathbf{X}_{[T]}^T \mathbf{X}_{[T]}]_{S,S}) > \lambda_{min}(\Sigma_{S,S})/2 > 0$.

If $K \leq 1$, we let $t = \sqrt{m}/(6C)$, $m\geq 16nC^2$. If $K > 1$, we let $t = \sqrt{m}/(6K^2C)$, $m\geq 16nK^4 C^2$. Under both cases, we have $K^2\max (\delta, \delta^2) = K^2 \delta < 1/2$. 

Therefore, if we have $T \in \Omega(k\log(p-k)/l)$ tasks, (\ref{step1_x}) holds with probability greater than $1-2e^{-C' Tl}$, where $C'$ is a constant.

\section{Bound of $\tilde{\mathbf{z}}_{S^c, 1}$ in Section \ref{sec:z_sc}}\label{app:bound1}
Recall that
$$
\tilde{\mathbf{z}}_{S^c, 1} = \mathbf{X}_{[T], S^c}^T \left\{\frac{1}{Tl}\mathbf{X}_{[T], S}(\hat\Sigma_{S,S})^{-1}\tilde{\mathbf{z}}_S + \Pi_{\mathbf{X}_{[T],S}^{\perp}} \left(\frac{\epsilon_{[T]}}{\lambda Tl}\right) \right\}
$$

In order to bound $\|\tilde{\mathbf{z}}_{S^c, 1}\|_\infty$, we consider each entry of the vector. That is, for $j\in S^c$, we need to bound
$$
\tilde{\mathbf{z}}_{j, 1} = \mathbf{X}_{[T], j}^T \left\{\frac{1}{Tl}\mathbf{X}_{[T], S}(\hat\Sigma_{S,S})^{-1}\tilde{\mathbf{z}}_S + \Pi_{\mathbf{X}_{[T],S}^{\perp}} \left(\frac{\epsilon_{[T]}}{\lambda Tl}\right) \right\}
$$

Since each entry in $\mathbf{X}_{[T], j} \in \mathbb{R}^n$ is sub-Gaussian, we know $\mathbf{X}_{[T], j}$ is also sub-Gaussian. We define $E_j^T$ by decomposition:
$$
\mathbf{X}_{[T], j} = \Sigma_{j,S}(\Sigma_{S,S})^{-1} \mathbf{X}_{[T], S} + E_{[T], j}^T
$$
where $E_{ij}^T$ is an independent sub-Gaussian random variable for all $i\in [T]$. We assume its variance proxy is $\sigma_e^2$ which is proportional to $\sigma_x^2$.

We can rewrite $\tilde{\mathbf{z}}_{j, 1}$ based on $E_{[T], j}$:
$$
\tilde{\mathbf{z}}_{j, 1} =   E_{[T], j} \underbrace{\left\{\frac{1}{Tl}\mathbf{X}_{[T], S}(\hat\Sigma_{S,S})^{-1}\tilde{\mathbf{z}}_S + \Pi_{\mathbf{X}_{[T],S}^{\perp}} \left(\frac{\epsilon_{[T]}}{\lambda Tl}\right) \right\}}_{A_j} + \underbrace{\Sigma_{j,S} (\Sigma_{S,S})^{-1} \tilde{\mathbf{z}}_S}_{B_j}
$$

By the mutual incoherence condition, we have $B_j = \Sigma_{j,S} (\Sigma_{S,S})^{-1} \tilde{\mathbf{z}}_S \leq 1-\gamma$. Therefore we only need to bound $\|A_j\|_2^2$ since the variance proxy for $E_{[T], j} A_j$ is $\|A_j\|_2^2\sigma_e^2$ (and we can bound $E_{[T], j} A_j$ by the concentration inequality of sub-Gaussian random variables).
$$
\|A_j\|_2^2 = A_j^T A_j = \frac{1}{Tl} \tilde{\mathbf{z}}_S^T (\hat\Sigma_{S,S})^{-1}\tilde{\mathbf{z}}_S + \left\|\Pi_{\mathbf{X}_{[T],S}^{\perp}} \left(\frac{\epsilon_{[T]}}{\lambda Tl}\right)\right\|_2^2
$$
For the first part $\frac{1}{Tl} \tilde{\mathbf{z}}_S^T (\hat\Sigma_{S,S})^{-1}\tilde{\mathbf{z}}_S$, by the techniques in appendix Section \ref{app:step1}, we have 
$$
\frac{1}{Tl} \tilde{\mathbf{z}}_S^T (\hat\Sigma_{S,S})^{-1}\tilde{\mathbf{z}}_S \leq \frac{1}{Tl} \|\tilde{\mathbf{z}}_S\|_2^2 (\lambda_{\min}(\hat\Sigma_{S,S}))^{-1} \leq \frac{1}{Tl}  \frac{2k}{\lambda_{min}(\Sigma_{S,S})}
$$
with probability $1-2e^{-C' Tl}$, where $C'$ is a constant.

For the second part, we have
$$
 \left\|\Pi_{\mathbf{X}_{[T],S}^{\perp}} \left(\frac{\epsilon_{[T]}}{\lambda Tl}\right)\right\|_2^2 \leq \frac{1}{\lambda^2 Tl}\frac{\|\epsilon_{[T]}\|_2^2}{Tl} = \frac{1}{\lambda^2 Tl} \frac{\sum_{i=1}^{Tl} \epsilon_{i}^2}{Tl} \leq \frac{C_1 \sigma_\epsilon^2}{\lambda^2 Tl}.
$$
By the concentration inequality of $\epsilon_i^2$ which is a sub-exponential random variable, we have that the last inequality holds with probability $1-2e^{-C_2 Tl}$ where $C_1, C_2$ are constants.

Now we define $M(T,l,k) := \sigma_e^2 \left(\frac{1}{Tl}  \frac{2k}{\lambda_{min}(\Sigma_{S,S})} + \frac{C_1\sigma_\epsilon^2}{\lambda^2 Tl}\right)$ and the event $\mathcal{T}_j = \left\{E_{[T], j} A_j > \gamma / 2  \right\}$. We have
$$
P\left(\bigcup_{j\in S^c}\mathcal{T}_j\right) \leq (p-k)\left(\mathrm{exp} \left(-\frac{\gamma^2}{8M(T,l,k)}\right) + 2e^{-C_3 Tl}\right)
$$

Therefore, if $\lambda\in {\Omega}\left(\sigma_\epsilon \sigma_x \sqrt{\frac{\log(p-k)}{Tl}}\right)$ and $T \in \Omega\left(\frac{k\log(p-k)}{l}\right)$, we have that $\tilde{\mathbf{z}}_{S^c, 1} < 1 - \gamma/2$ holds with probability greater than $1-2e^{-C_4 \log(p-k)}$.

}

\section{Bound of estimation error}\label{app:bound_err}
The second part in Theorem \ref{theorem_meta} is about the estimation error. We first write the estimation error in the following form:
$$
\begin{aligned}
    \tilde{\mathbf{w}}_S - {\mathbf{w}}_S^* &= \hat\Sigma_{S,S}^{-1}  \left(\frac{1}{Tl}\sum_{i=1}^T \mathbf{X}_{t_i, S}^T\epsilon_{t_i} - \lambda\tilde{\mathbf{z}}_S + \frac{1}{Tl}\sum_{i=1}^T \mathbf{X}_{t_i, S}^T\mathbf{X}_{t_i, S} \Delta_{t_i, S}^* \right) \\
     &= \underbrace{\hat\Sigma_{S,S}^{-1} \frac{1}{Tl}\sum_{i=1}^T \mathbf{X}_{t_i, S}^T\epsilon_{t_i}}_{F_1}  - \underbrace{\hat\Sigma_{S,S}^{-1}\lambda\tilde{\mathbf{z}}_S}_{F_2}    +  \underbrace{\hat\Sigma_{S,S}^{-1}\lambda \frac{1}{\lambda Tl}\sum_{i=1}^T \mathbf{X}_{t_i, S}^T\mathbf{X}_{t_i, S} \Delta_{t_i, S}^*}_{F_3} 
\end{aligned}
$$

{
By the technique in appendix Section \ref{app:step1}, we know 
$$
\|F_2\|_\infty \leq \lambda \sqrt{k} \frac{2}{\lambda_{min}(\Sigma_{S,S})}
$$
holds with probability greater than $1-2e^{-C' Tl}$, where $C'$ is a constant.

For $j\in S$, we have
$$
\frac{1}{Tl}\sum_{i=1}^T \mathbf{X}_{t_i, j}^T\epsilon_{t_i} = \frac{1}{Tl}\sum_{i=1}^T \sum_{m=1}^l {X}_{t_i, j, m}\epsilon_{t_i, m}
$$
which can be bounded by the concentration inequality of sub-exponential random variables. Here we let $\|{X}_{t_i, j, m}\epsilon_{t_i, m}\|_{\psi_1} = M$. By Lemma \ref{orlicz_prod_sum}, we know $M \in O(\sigma_x \sigma_\epsilon)$. We then use Theorem 1.4 in \cite{gotze2019concentration} again:
$$
P\left(\left|\frac{1}{Tl}\sum_{i=1}^T \mathbf{X}_{t_i, j}^T\epsilon_{t_i} \right| \leq t \right) \geq 1-2 \mathrm{exp}\left(\frac{1}{C} \min\left(\frac{t^2 Tl}{M^2}, \frac{tTl}{M}\right)\right)
$$
We let $t=\lambda$, and $T \in \Omega\left(\frac{k\log(p-k)}{l}\right)$, then 
$$
\|F_1\|_\infty \leq \lambda \sqrt{k} \frac{2}{\lambda_{min}(\Sigma_{S,S})}
$$
holds with probability greater than $1-2e^{-c_5 k\log(p-k) }$, where $C_5$ is a constant.

For $F_3$, we use the definition in (\ref{eq:zeta_S}) and we set $\gamma = 1$ in the bound (\ref{eq:bound_zeta_S}). We know that
$$
\|F_3\|_\infty \leq \lambda \sqrt{k} \frac{2}{\lambda_{min}(\Sigma_{S,S})}
$$
holds with probability greater than $1-c_7 e^{-c_{7} \log(p-k)}$.

Therefore, we can bound the estimation error: with probability greater than $1-c_8e^{-c_9 \log(p-k)}$, we have
$$
\|\tilde{\mathbf{w}}_S - {\mathbf{w}}_S^*\|_\infty \leq  \frac{6\lambda \sqrt{k}}{\lambda_{min}(\Sigma_{S,S})}
$$

\section{Proof of Theorem \ref{theorem_new}}\label{app:sec_proof_novel}
We use the primal dual witness framework as in the proof of Theorem \ref{theorem_meta}. Since for this novel $(T+1)$-th task, $\Delta_{t_i}^*, i=1,2,\cdots, T$ is not considered, the choice of $l$ and $\lambda$ can be more flexible. We set $l\in\Omega(k'\log(k-k'))$ and $\lambda \in \Omega(\sqrt{\log(k-k')/l})$.

For \textbf{step 1}, similar to the \textbf{step 1} in Theorem \ref{theorem_meta}, with probability greater than $1-2e^{-C'l}$, we have
$$
\frac{1}{l}[\mathbf{X}_{T+1}^T \mathbf{X}_{T+1}]_{S,S} \succ 0
$$
For \textbf{step 5}, we only have one part which is $\tilde{\mathbf{z}}_{S^c, 1}$ in the proof of \textbf{step 5} in Theorem \ref{theorem_meta}. We can use the technique in appendix Section \ref{app:bound1}. With probability greater than $1-2e^{-c_4 \log(k-k')}$, we have
$$
\tilde{\mathbf{z}}_{S^c, 1, T+1} \leq 1 - \gamma/2.
$$
For the estimation error bound, we can use the technique in appendix Section \ref{app:bound_err}. We only need to consider the two parts $F_1, F_2$ which does not contain $\Delta_{t_i}^*$. With probability greater than $1-c_8 e^{-c_9 \log(k-k')}$, we have 
$$
\|\hat{\mathbf{w}}_{T+1} - (\mathbf{w}^* + \Delta^*_{t_{T+1}})\|_\infty \leq  \frac{4\lambda \sqrt{k'}}{\lambda_{min}(\Sigma_{S,S})}.
$$
}

\section{Proof of Theorem \ref{theorem_fano}}\label{app:sec_proof_fano}

We first introduce Fano's inequality \cite{fano1952class, Yu97} (the version below can also be found directly in \cite{scarlett2019introductory}).

\begin{Lemma}\label{fano_ineq}(Fano's inequality)
With input dataset $S$, for any estimator $\hat{\theta}(S)$ with $k$ possible outcomes, i.e., $\hat\theta \in \Theta, |\Theta|=k$, if $S$ is generated from a model with true parameter $\theta^*$ chosen uniformly at random from the same $k$ possible outcomes $\Theta$, we have:
$$
\mathbb{P}[\hat{\theta}(S) \neq \theta^*] \geq 1 - \frac{\mathbb{I}(\theta^*, S) + \log 2}{\log k}
$$
\end{Lemma}

Now we show that $\mathbb{I}(\theta^*, S) \leq Tl\cdot c_1 + l_{T+1}\cdot c_2$, where $c_1, c_2$ are constants, and $\theta^*$ represents the parameter $(\mathbf{w}^*, \Delta^*_{t_{T+1}})$ we want to recover. Here $S$ is all the data in the $T+1$ tasks, $S_{[T]}$ is the data in the first $T$ tasks, and $S_i$ is the data of task $t_{i}$. The mutual information is bounded by the following steps.

\begin{align*}
    \mathbb{I}(\theta^*, S) &= \frac{1}{k}\sum_{\theta^* \in \Theta} \int_{S} p_{S | \theta^*} (S) \log\frac{p_{S | \theta^*} (S)}{p_{S}(S)} d S = \frac{1}{k} \sum_{\theta^* \in \Theta} \int_{S} p_{S | \theta^*} (S) \log\frac{p_{S | \theta^*} (S)}{\frac{1}{k}\sum_{\theta' \in \Theta}p_{S|\theta'}(S)} d S \\
    &\leq \frac{1}{k^2} \sum_{\theta^* \in \Theta}\sum_{\theta' \in \Theta} \int_{S} p_{S| \theta^*}(S) \log \frac{p_{S| \theta^*}(S)}{p_{S| \theta'}(S)} dS =\frac{1}{k^2} \sum_{\theta^* \in \Theta}\sum_{\theta' \in \Theta} \mathbb{KL}(P_{S| \theta^*}||P_{S| \theta'})
\end{align*}

Given the common coefficient $w^*$, the data for each task is independent from each other. Therefore we have
\begin{align}\label{KL}
\mathbb{KL}(P_{S| \theta^*}||P_{S| \theta'}) &= \mathbb{KL}(P_{S_{[T]}| \theta^*}||P_{S_{[T]}| \theta'}) + \mathbb{KL}(P_{S_{T+1}| \theta^*}||P_{S_{T+1}| \theta'}) 
\end{align}

First, we consider the first part (\ref{KL}). We use $S'$ to denote $S_{[T]}$. Let $P_{S'}=P_{S_{[T]}|\theta*}, P'_{S'}=P_{S_{[T]}|\theta'}$. Note that

$$
\mathbb{KL}(P_{S'}||P'_{S'}) = \int_{S'} P_{S'}\log\frac{P_{S'}}{P'_{S'}} dS'
$$

Furthermore
\begin{align*}
P_{S'} &= \int_{\Delta^*_{t_1},\cdots,\Delta^*_{t_T}} P_{S'|\mathbf{w}^*,\Delta^*_{t_1},\cdots,\Delta^*_{t_T}} d \Delta^*_{t_1},\cdots,d\Delta^*_{t_T} = \int_{\Delta^*_{t_1}} P_{S_1|\mathbf{w}^*,\Delta^*_{t_1}} d \Delta^*_{t_1} \cdots \int_{\Delta^*_{t_T}} P_{S_T|\mathbf{w}^*,\Delta^*_{t_T}} d \Delta^*_{t_T}
\end{align*}

This is because conditioning on $\mathbf{w}^*,\Delta^*_{t_1},\cdots,\Delta^*_{t_T}$, the data for each task is independent and therefore
\begin{align*}
P_{S'|\mathbf{w}^*,\Delta^*_{t_1},\cdots,\Delta^*_{t_T}} &= P_{S_1|\mathbf{w}^*,\Delta^*_{t_1},\cdots,\Delta^*_{t_T}}\cdots P_{S_T|\mathbf{w}^*,\Delta^*_{t_1},\cdots,\Delta^*_{t_T}} = P_{S_1|\mathbf{w}^*,\Delta^*_{t_1}}P_{S_2|\mathbf{w}^*,\Delta^*_{t_2}}\cdots P_{S_T|\mathbf{w}^*,\Delta^*_{t_T}} 
\end{align*}

If we set $a_i = P_{S_i|\mathbf{w}^*,\Delta^*_{t_i}}$, $a'_i = P_{S_i|\mathbf{w}',\Delta'_{t_i}}$, we have
$$
P_{S'} = a_1 a_2\cdots a_T,\quad P'_{S'} = a'_1 a'_2\cdots a'_T. 
$$

Therefore 
$$
\mathbb{KL}(P_{S'}||P'_{S'}) = \int_{S'} a_1\cdots a_T \left(\log\frac{a_1}{a'_1} + \cdots + \log\frac{a_T}{a'_T}\right) dS'
$$

We know $a_i$ is a function of $S_j$ only when $i=j$, and $\int_{S_j} a_j\ dS_j = 1$. Therefore, we have  
$$
\int_{S'} a_1 a_2\cdots a_T \left(\log\frac{a_i}{a'_i}  \right) dS' = \int_{S_i} a_i \log\frac{a_i}{a'_i}  dS_i 
$$

Therefore
\begin{align*}
\mathbb{KL}(P_{S'}||P'_{S'}) &= \sum_{i=1}^T \int_{S_i} a_i \log\frac{a_i}{a'_i}  dS_i \leq T \max_{i\in\{1,2,\cdots,T\}}\int_{S_i} a_i \log\frac{a_i}{a'_i}  dS_i 
\end{align*}
For any task $t_i$, conditioning on $(\mathbf{w}^*, \Delta^*_{t_i})$, we know all samples in $S_i$ are i.i.d. If we set $S_{i,j}$ to be the $j$-th sample in the task $t_i$, and $a_{i,j} = P_{S_{i,j}|\mathbf{w}^*,\Delta^*_{t_i}}$, we have
$$
\int_{S_i} a_i \log\frac{a_i}{a'_i}  dS_1 = l \int_{S_{i,1}} a_{i,1} \log\frac{a_{i,1}}{a'_{i,1}}  dS_{i,1}
$$

Therefore,
\begin{align*}
\mathbb{KL}(P_{S'}||P'_{S'}) &\leq Tl \max_i \int_{S_{i,1}} P_{S_{i,1}|\mathbf{w}^*, \Delta^*_{t_i}} \frac{P_{S_{i,1}|\mathbf{w}^*, \Delta^*_{t_i}}}{P_{S_{i,1}|\mathbf{w}', \Delta'_{t_i}}} dS_{i,1} = Tl\cdot c_1
\end{align*}

Then we consider the second part (\ref{KL}). For the task $t_{T+1}$, conditioning on $(\mathbf{w}^*, \Delta^*_{t_{T+1}})$, since we know all samples in $S_{T+1}$ are i.i.d., we have
\begin{align*}
&\mathbb{KL}(P_{S_{T+1}| \theta^*}||P_{S_{T+1}| \theta'}) = l_{T+1} \int_{S_{T+1,1}} P_{S_{T+1,1}|\mathbf{w}^*, \Delta^*_{t_{T+1}}} \frac{P_{S_{T+1,1}|\mathbf{w}^*, \Delta^*_{t_{T+1}}}}{P_{S_{T+1,1}|\mathbf{w}', \Delta'_{t_{T+1}}}} dS_{T+1,1} = l_{T+1} \cdot c_2
\end{align*}

Combining the results above, we have
$$
\mathbb{I}(\theta^*, S) \leq Tl\cdot c_1 + l_{T+1}\cdot c_2.
$$

Finally, from Fano's inequality, we know
$$
\mathbb{P}[\hat\theta \neq \theta^*] \geq 1 - \frac{\log 2 + Tl\cdot c_1 + l_{T+1}\cdot c_2}{\log |\Theta|}
$$

\section{Additional experiments} \label{app:experiment}

In this section, we first present simulations to show that Theorem \ref{theorem_meta} holds in the sense that for different choices of $l$ and $p$, one only needs $T = c \cdot \left(k\log(p-k)/l\right)$ to recover the true common support $S$ with high probability. 
We then perform a real-world experiment with a gene expression dataset from \cite{kouno2013temporal} which was used in the experimental validation of \cite{ollier2017regression}.
Our meta-learning method has lower mean square error (MSE) of the prediction on a new task than multi-task methods.

\subsection{Simulations} 

For all the experiments in this section, we let $k=|S|=5$, and perform $100$ repetitions for each setting. We compute the \emph{empirical} probability of successful support recovery $P(\hat{S}=S)$ as the number of times we obtain exact support recovery among the $100$ repetitions, divided by $100$. 
We compute the standard deviation as $\sqrt{P(\hat{S}=S)(1-P(\hat{S}=S))/100}$, that is, by using the formula of the standard deviation of the Binomial distribution.
For the estimation error $\|\hat{\mathbf{w}} - \mathbf{w}^*\|_\infty$, we calculate the mean and standard deviation by using the empirical results of the $100$ repetitions.

\subsubsection{Gaussian distribution setting} \label{sec:app_gaussian}

\begin{figure}[!htbp]
    \centering
    \includegraphics[width=0.75\textwidth]{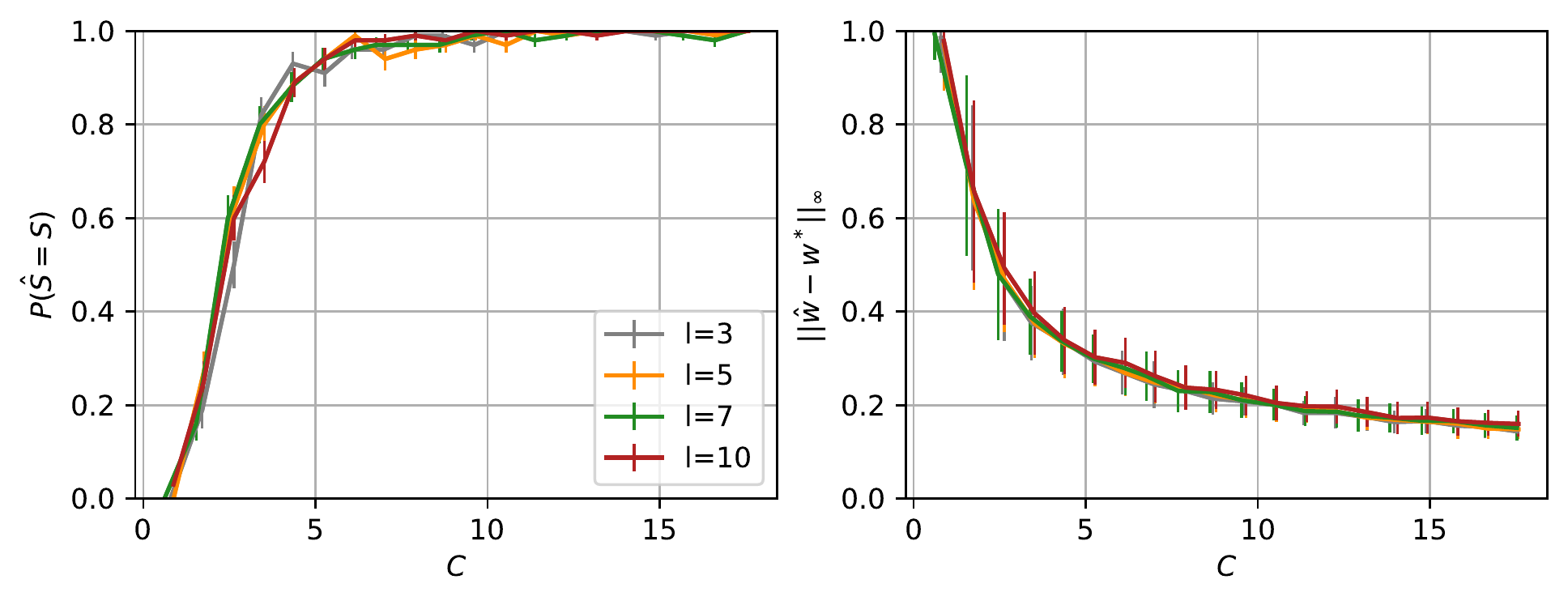}
    \caption{
       Simulations with our meta sparse regression under Gaussian distributions of $\epsilon_{t_i,j}, \Delta_{t_i,m}^*, X_{t_i,j,m}\ \forall i\in[T],j\in[l],m\in S$. We use $\lambda=\sqrt{\frac{\log p }{Tl}}$.
       \textbf{Left:} Probability of exact support recovery for different number of tasks under various settings of sample size $l$. The x-axis is set by $C:=\frac{Tl}{k\log(p-k)}$.  \textbf{Right:} The corresponding estimation error of the common parameter $\mathbf{w}$ in $\ell_\infty$ norm. 
    } \label{fig:gaussian_change_l}
\end{figure}

We first consider the setting of different sample size $l$. We choose $l\in\{3,5,7,10\}$ and use $\lambda=\sqrt{\log p/(Tl)}$ for all the pairs of $(T,l)$. We denote the set $\{1,2,3,\cdots,a\}$ by $[a]$. For all $i\in[T],j\in[l],m\in S$, we set $\epsilon_{t_i,j} \sim N(\mu=0,\sigma_\epsilon=0.1)$, $\Delta_{t_i,m}^* \sim N(\mu=0,\sigma_\Delta=0.2)$, $X_{t_i,j,m} \sim N(\mu=0,\sigma_x=1)$, which are mutually independent. We set $p=100$, and $\mathbf{w}^*$ having five entries equal to 1, and the rest of the entries being 0. The support of $\Delta_{t_i}^*$ is same as the support of $\mathbf{w}^*$. The results are shown in Figure \ref{fig:gaussian_change_l}. The number of tasks $T$ is rescaled to $C$ defined by $\frac{Tl}{k\log(p-k)}$. For different choices of $l$, the curves overlap with each other perfectly (for both $P(\hat{S}=S)$ and $\|\hat{\mathbf{w}} - \mathbf{w}^*\|_\infty$). 

\begin{figure}[!htbp]
    \centering
    \includegraphics[width=0.75\textwidth]{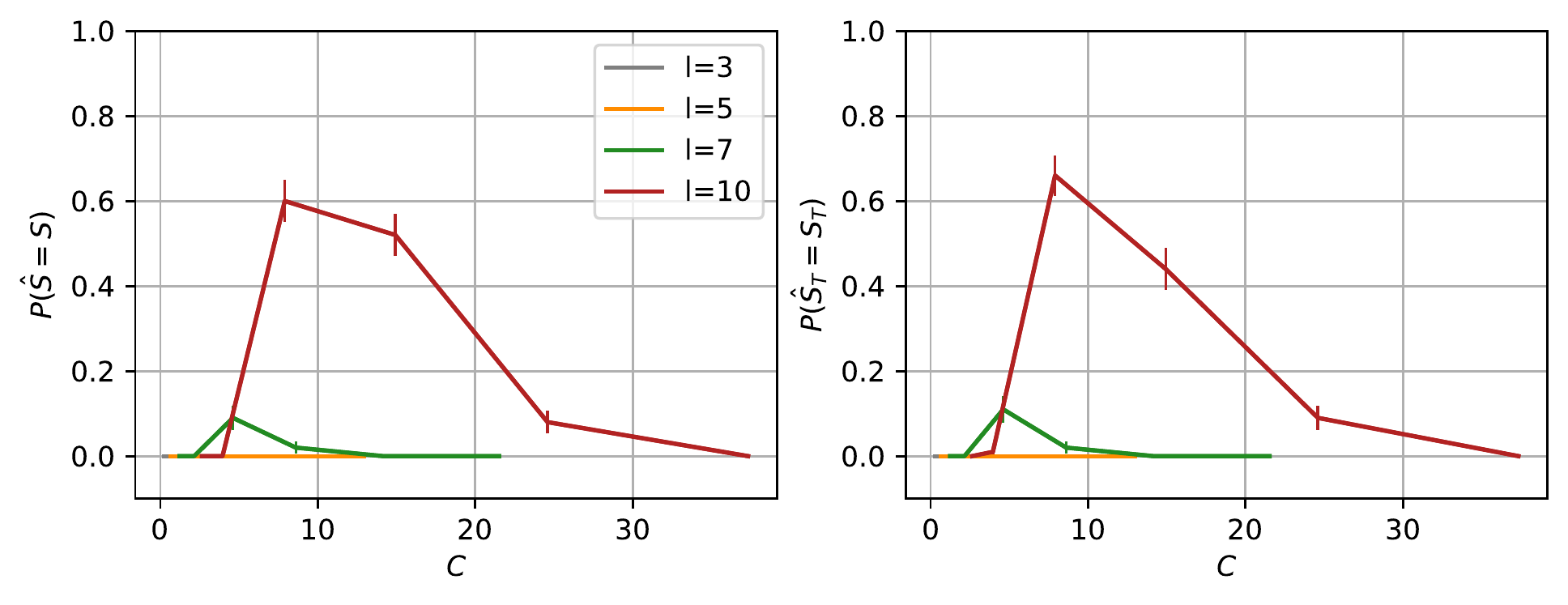}
    \caption{
       Simulations with the multi-task method with ${\ell_{1,2}}$ regularization under Gaussian distributions of $\epsilon_{t_i,j}, \Delta_{t_i,m}^*, X_{t_i,j,m}\ \forall i\in[T],j\in[l],m\in S$. We use $\lambda_{1,2}=30\sqrt{\log p/(Tl)}$.
       \textbf{Left:} Probability of exact support union recovery ($S=\hat{S}:=\bigcup_{i=1}^T \hat{S}_i$) for different number of tasks under various settings of sample size $l$. The x-axis is set by $C:=\frac{Tl}{k\log(p-k)}$.  \textbf{Right:} Probability of exact support recovery of the last task ($\hat{S}_{T}=S_T$). 
    } \label{fig:gaussian_change_l_grouplasso}
\end{figure}

\begin{figure}[!htbp]
    \centering
    \includegraphics[width=0.75\textwidth]{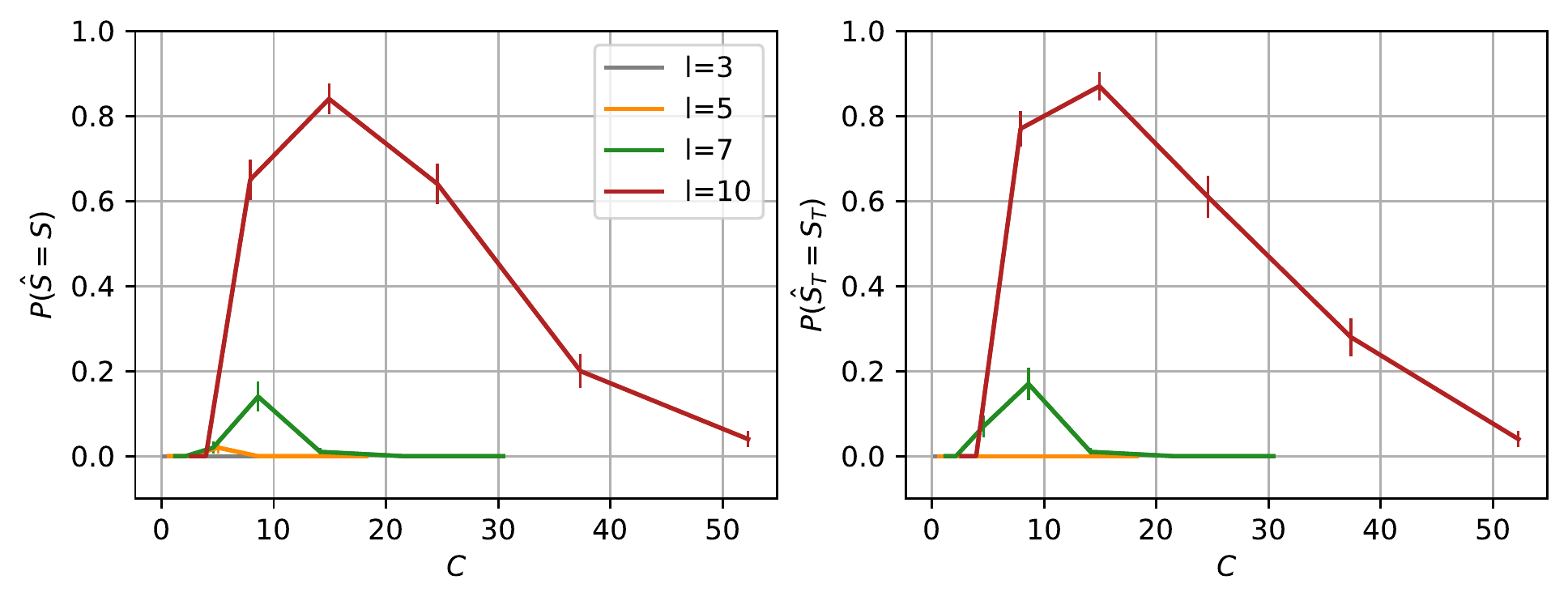}
    \caption{
       Simulations with the multi-task method with $\ell_1+\ell_{1,\infty}$ regularization under Gaussian distributions of $\epsilon_{t_i,j}, \Delta_{t_i,m}^*, X_{t_i,j,m} \ \forall i\in[T],j\in[l],m\in S$. We use $\lambda_1=30\sqrt{\log p/(Tl)}, \lambda_{1,\infty}=(1+1.5 T)\lambda_1 /2.5$.
       \textbf{Left:} Probability of exact support union recovery ($S=\hat{S}:=\bigcup_{i=1}^T \hat{S}_i$) for different number of tasks under various settings of sample size $l$. The x-axis is set by $C:=\frac{Tl}{k\log(p-k)}$.  \textbf{Right:} Probability of exact support recovery of the last task ($\hat{S}_{T}=S_T$). 
    } \label{fig:gaussian_change_l_dirtymodel}
\end{figure}

Next we show that the problem described above cannot be solved by multi-task methods. We use two multi-task methods with regularization terms being ${\ell_{1,2}}$ \cite{obozinski2011support} and $\ell_1+\ell_{1,\infty}$ \cite{jalali2010dirty} respectively. The results are shown in Figure \ref{fig:gaussian_change_l_grouplasso} and \ref{fig:gaussian_change_l_dirtymodel}, where we take $\hat{S}=\bigcup_{i=1}^T \hat{S}_i$. We show both $P(\hat{S}=S)$ and $P(\hat{S}_{T}=S_T)$ since the multi-task learning methods are not designed for recovering only the union of the supports of all tasks. As we claimed in Table \ref{tab:comparison}, the multi-task methods require that $l$ grows with $T$ in order to retain the probability of support recovery. Therefore we see when $l$ is fixed at $3,5,7,10$, the probability of support recovery first increases then decreases to $0$ as $T$ increases. For the ${\ell_{1,2}}$ method of \cite{obozinski2011support}, we use $\lambda_{1,2}=30\sqrt{\log p/(Tl)}$ as the parameter for the $\ell_{1,2}$ norm; for 
the $\ell_1+\ell_{1,\infty}$ method of \cite{jalali2010dirty}, we use $\lambda_1=30\sqrt{\log p/(Tl)}$ as the parameter of the $\ell_1$ norm and $\lambda_{1,\infty}=(1+1.5 T)\lambda_1 /2.5$ as the parameter of the $\ell_{1,\infty}$ norm. We also tried different choices of $\lambda_{1,2}, \lambda_1, \lambda_{1,\infty}$ and the trends of the results are similar.

The Figure \ref{fig:simulation} is from the results in Figure \ref{fig:gaussian_change_l}, \ref{fig:gaussian_change_l_grouplasso}, \ref{fig:gaussian_change_l_dirtymodel}.

\begin{figure}[!htbp]
    \centering
    \includegraphics[width=0.75\textwidth]{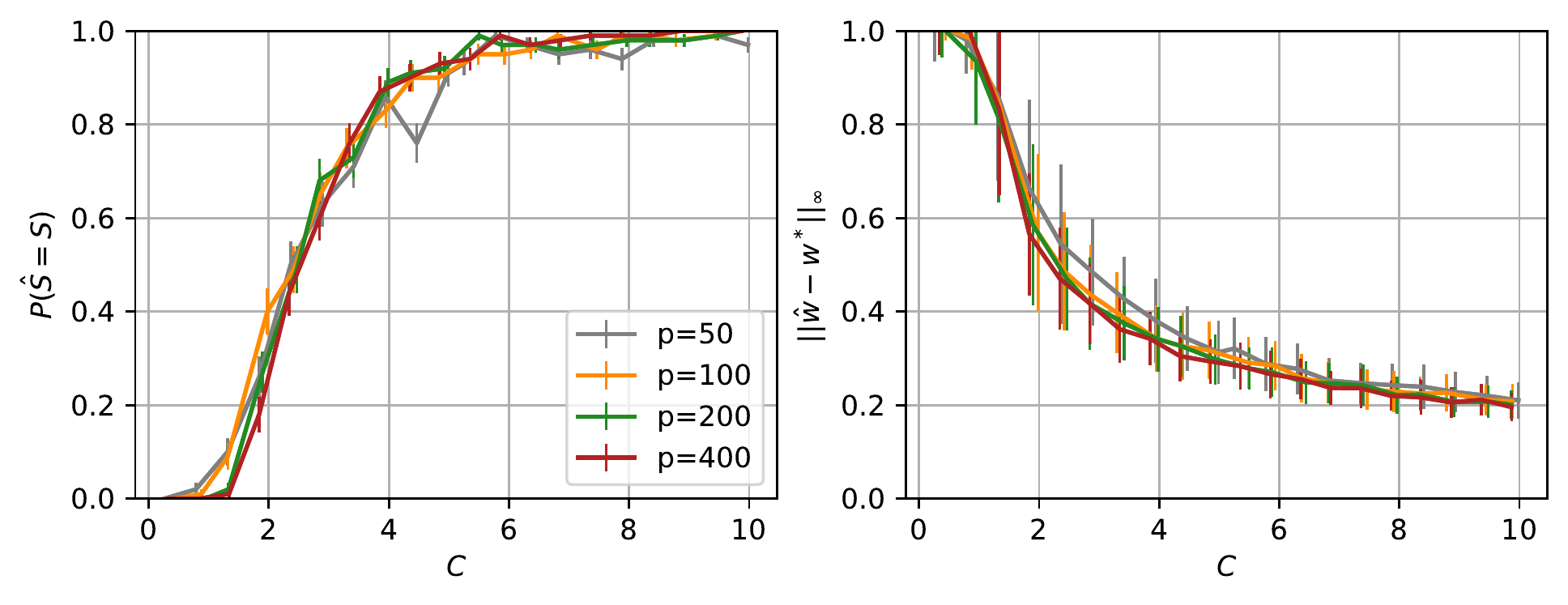}
    \caption{
        Simulations with our meta sparse regression under Gaussian distributions of $\epsilon_{t_i,j}, \Delta_{t_i,m}^*, X_{t_i,j,m}\ \forall i\in[T],j\in[l],m\in S$. We use $\lambda=\sqrt{\frac{\log p }{Tl}}$.
        \textbf{Left:} Probability of exact support recovery for different number of tasks under various settings of number of parameters $p$. The x-axis is set by $C:=\frac{Tl}{k\log(p-k)}$.  \textbf{Right:} The corresponding estimation error of the common parameter $\mathbf{w}$ in $\ell_\infty$ norm. 
    } \label{fig:gaussian_change_p}
\end{figure}

Then, for our method we consider the setting of different number of parameters $p$. We choose $p\in\{50,100,200,400\}$ and use $\lambda=\sqrt{\log p/(Tl)}$ for all the pairs of $(T,l)$. For all $i\in[T],j\in[l],m\in S$, we set $\epsilon_{t_i,j} \sim N(\mu=0,\sigma_\epsilon=0.1)$, $\Delta_{t_i,m}^* \sim N(\mu=0,\sigma_\Delta=0.2)$, $X_{t_i,j,m} \sim N(\mu=0,\sigma_x=1)$, which are mutually independent. We set $l=5$, and $\mathbf{w}^*$ having five entries equal to 1, and the rest of the entries being 0. The support of $\Delta_{t_i}^*$ is same as the support of $\mathbf{w}^*$. The results are shown in Figure \ref{fig:gaussian_change_p}. The number of tasks $T$ is rescaled to $C$ defined by $\frac{Tl}{k\log(p-k)}$. For different choices of $p$, the curves overlap with each other perfectly (for both $P(\hat{S}=S)$ and $\|\hat{\mathbf{w}} - \mathbf{w}^*\|_\infty$).

\subsubsection{Uniform distribution setting} 

In this paper we only assume that the distributions are sub-Gaussian which includes the uniform distribution. Therefore in this section, we replace the Gaussian distribution setting in the appendix Section \ref{sec:app_gaussian} with a uniform distribution setting.

\begin{figure}[!htbp]
    \centering
    \includegraphics[width=0.75\textwidth]{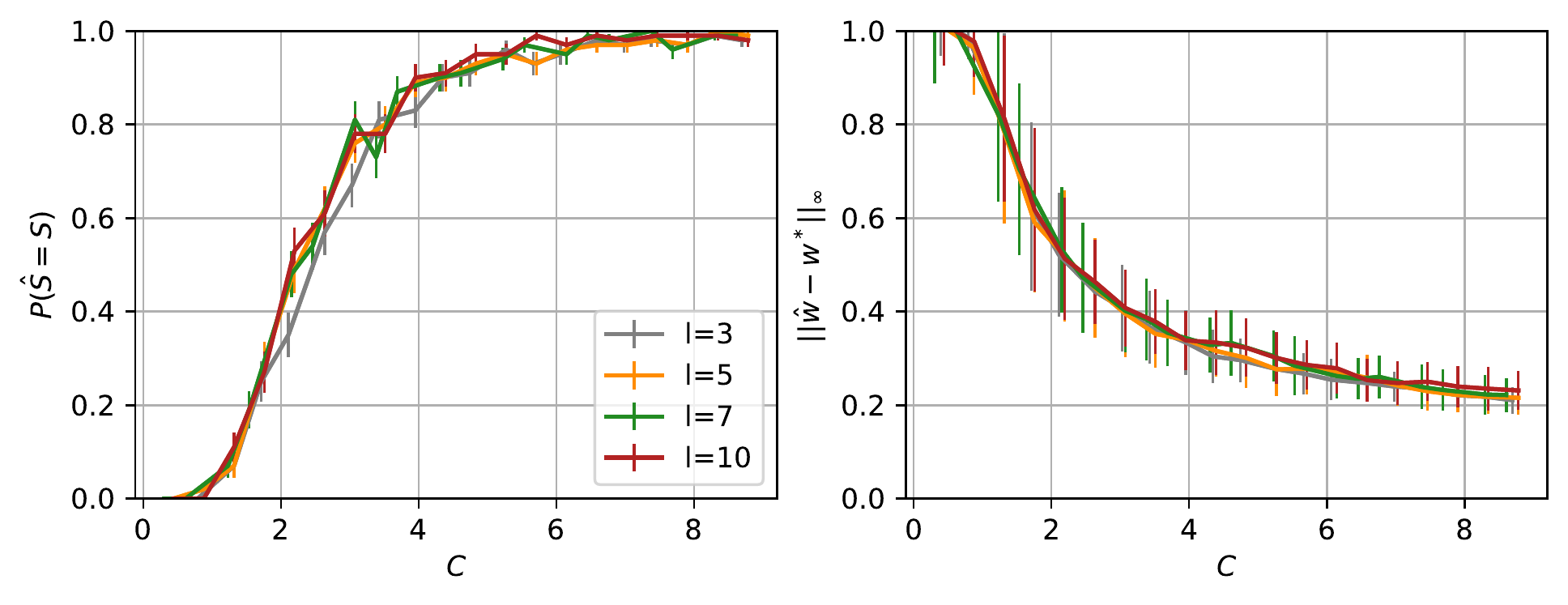}
    \caption{
        Simulations with our meta sparse regression under uniform distributions of $\epsilon_{t_i,j}, \Delta_{t_i,m}^*, X_{t_i,j,m}\ \forall i\in[T],j\in[l],m\in S$. We use $\lambda=\sqrt{\frac{\log p }{Tl}}$.
        \textbf{Left:} Probability of exact support recovery for different number of tasks under various settings of $l$. The x-axis is set by $C:=\frac{Tl}{k\log(p-k)}$.  \textbf{Right:} The corresponding estimation error of the common parameter $\mathbf{w}$ in $\ell_\infty$ norm. 
    } \label{fig:uniform_change_l}
\end{figure}

For all $i\in[T],j\in[l],m\in S$, we set $\epsilon_{t_i,j} \sim \text{Uniform}(-0.1\sqrt{3},0.1\sqrt{3})$, $\Delta_{t_i,m}^* \sim \text{Uniform}(-0.2\sqrt{3},0.2\sqrt{3})$, $X_{t_i,j,m} \sim \text{Uniform}(-\sqrt{3},\sqrt{3})$, which are mutually independent. 
We consider the setting of different sample size $l$. We choose $l\in\{3,5,7,10\}$ and use $\lambda=\sqrt{\log p/(Tl)}$ for all the pairs of $(T,l)$. 
We set $p=100$, and $\mathbf{w}^*$ having five entries equal to 1, and the rest of the entries being 0. The support of $\Delta_{t_i}^*$ is same as the support of $\mathbf{w}^*$. The results are shown in Figure \ref{fig:uniform_change_l}. The number of tasks $T$ is rescaled to $C$ defined by $\frac{Tl}{k\log(p-k)}$. For different choices of $l$, the curves overlap with each other perfectly (for both $P(\hat{S}=S)$ and $\|\hat{\mathbf{w}} - \mathbf{w}^*\|_\infty$). 

\begin{figure}[!htbp]
    \centering
    \includegraphics[width=0.75\textwidth]{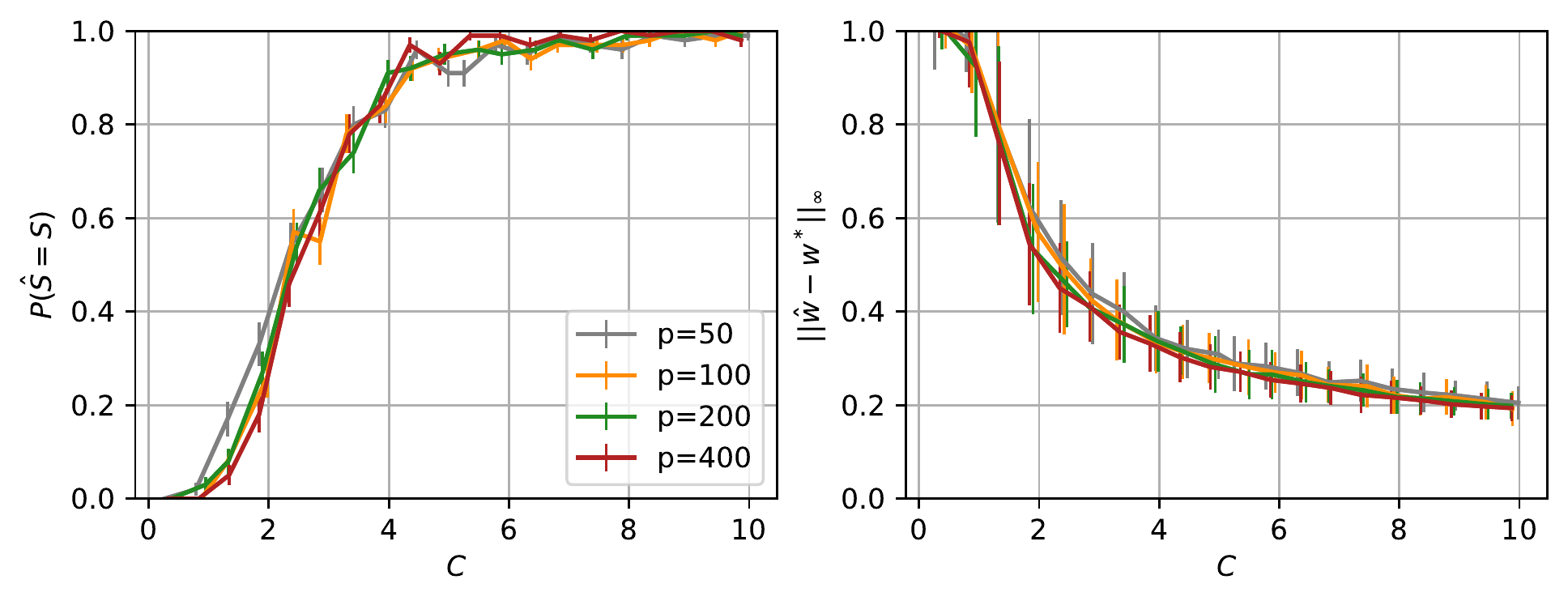}
    \caption{
        Simulations with our meta sparse regression under uniform distributions of $\epsilon_{t_i,j}, \Delta_{t_i,m}^*, X_{t_i,j,m}\ \forall i\in[T],j\in[l],m\in S$. We use $\lambda=\sqrt{\frac{\log p }{Tl}}$.
        \textbf{Left:} Probability of exact support recovery for different number of tasks under various settings of number of parameters $p$. The x-axis is set by $C:=\frac{Tl}{k\log(p-k)}$.  \textbf{Right:} The corresponding estimation error of the common parameter $\mathbf{w}$ in $\ell_\infty$ norm. 
    } \label{fig:uniform_change_p}
\end{figure}

Then we consider the setting of different number of parameters $p$. We choose $p\in\{50,100,200,400\}$ and use $\lambda=\sqrt{\log p/(Tl)}$ for all the pairs of $(T,l)$. For all $i\in[T],j\in[l],m\in S$, we set $\epsilon_{t_i,j} \sim \text{Uniform}(-0.1\sqrt{3},0.1\sqrt{3})$, $\Delta_{t_i}^* \sim \text{Uniform}(-0.2\sqrt{3},0.2\sqrt{3})$, $X_{t_i,j,m} \sim \text{Uniform}(-\sqrt{3},\sqrt{3})$, which are mutually independent. We set $l=5$, and $\mathbf{w}^*$ having five entries equal to 1, and the rest of the entries being 0. The results are shown in Figure \ref{fig:uniform_change_p}. The number of tasks $T$ is rescaled to $C$ defined by $\frac{Tl}{k\log(p-k)}$. For different choices of $p$, the curves overlap with each other perfectly (for both $P(\hat{S}=S)$ and $\|\hat{\mathbf{w}} - \mathbf{w}^*\|_\infty$).

\subsubsection{Mixture of sub-Gaussian distribution setting} 

In Section \ref{sec:problem_setting}, we state that we can consider the setting $S_i \subseteq S$ under the sub-Gaussian distribution assumption. Therefore in this section, we replace the Gaussian distribution setting of $\Delta_{t_i,m}^*$ in the appendix Section \ref{sec:app_gaussian} with a mixture of sub-Gaussian distribution setting. More specifically, we consider a mixture of a Dirac distribution and a Gaussian distribution.

For all $i\in[T],j\in[l],m\in S$, we set $\epsilon_{t_i,j} \sim N(\mu=0,\sigma_\epsilon=0.1)$, $X_{t_i,j,m} \sim N(\mu=0,\sigma_x=1)$, $\Delta_{t_i,m}^* \sim 0.5\  \delta_{-\mathbf{w}^*_m} + 0.5\ N(\mu=0,\sigma_\Delta=0.2)$, which are mutually independent.
We consider the setting of different sample size $l$. We choose $l\in\{3,5,7,10\}$ and use $\lambda=4\sqrt{\log p/(Tl)}$ for all the pairs of $(T,l)$. 
We set $p=100$, and $\mathbf{w}^*$ having five entries equal to 2, and the rest of the entries being 0. The support of $\Delta_{t_i}^*$ is same as the support of $\mathbf{w}^*$ denoted by $S$ while the support of $\mathbf{w}^* + \Delta_{t_i}^*$ could be a subset of $S$, i.e., $S_i \subseteq S$. More specifically, the distribution of $\Delta_{t_i,m}^*$ means that for the $m$-th parameter in the $i$-th task, i.e., $w_{i,m} := [\mathbf{w}^* + \Delta_{t_i}^*]_m, \forall i\in[T], m\in S$, there is a 50\% probability that $w_{i,m} = 0$, and a 50\% probability that $w_{i,m} \in N(2,0.2)$. 

\begin{figure}[!htbp]
    \centering
    \includegraphics[width=0.75\textwidth]{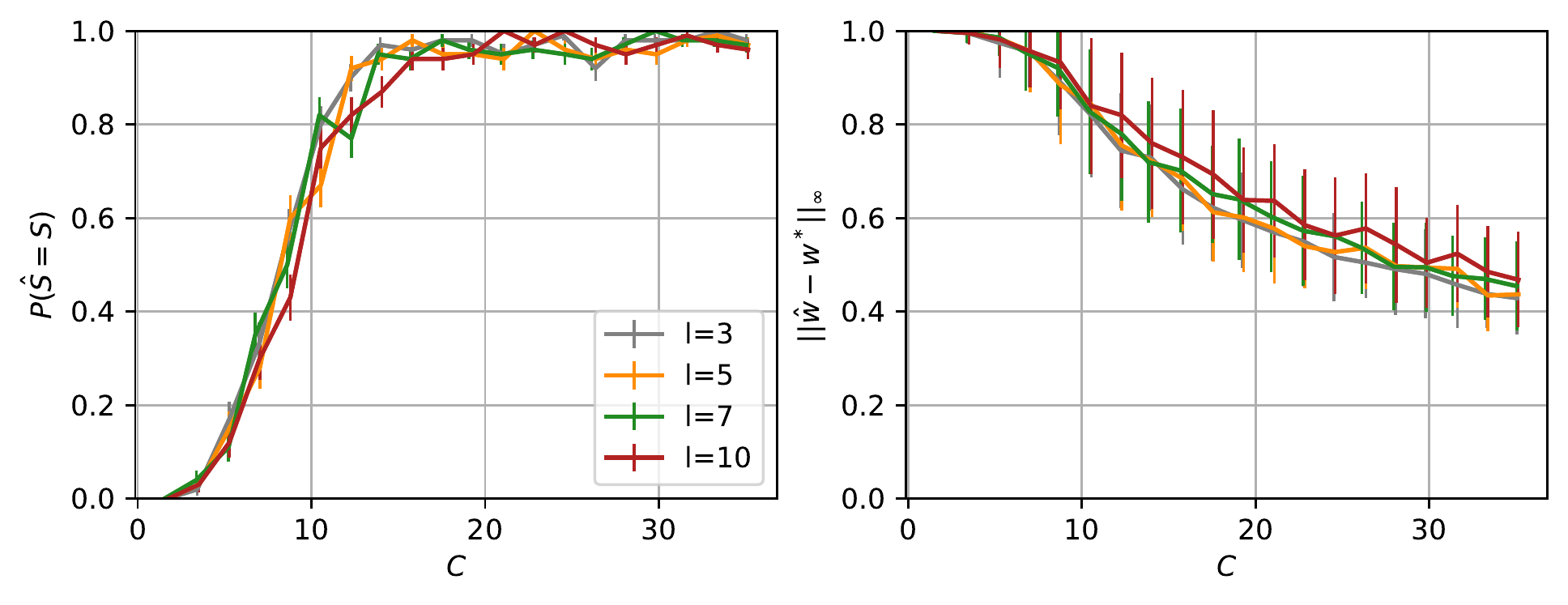}
    \caption{
        Simulations with our meta sparse regression under Gaussian distributions of $\epsilon_{t_i,j}, X_{t_i,j,m}\ \forall i\in[T],j\in[l],m\in S$ and a mixture of sub-Gaussian distributions of $\Delta_{t_i,m}^*\ \forall i\in[T],m\in S$ such that the support of $\mathbf{w}^* + \Delta_{t_i}^*$ could be a subset of $S$, i.e., $S_i \subseteq S$. We use $\lambda=4\sqrt{\frac{\log p }{Tl}}$.
        \textbf{Left:} Probability of exact support recovery for different number of tasks under various settings of $l$. The x-axis is set by $C:=\frac{Tl}{k\log(p-k)}$.  \textbf{Right:} The corresponding estimation error of the common parameter $\mathbf{w}$ in $\ell_\infty$ norm. 
    } \label{fig:gaussian_mix_change_l}
\end{figure}

The results are shown in Figure \ref{fig:gaussian_mix_change_l}. The number of tasks $T$ is rescaled to $C$ defined by $\frac{Tl}{k\log(p-k)}$. For different choices of $l$, the curves overlap with each other perfectly (for both $P(\hat{S}=S)$ and $\|\hat{\mathbf{w}} - \mathbf{w}^*\|_\infty$).

\begin{figure}[!htbp]
    \centering
    \includegraphics[width=0.75\textwidth]{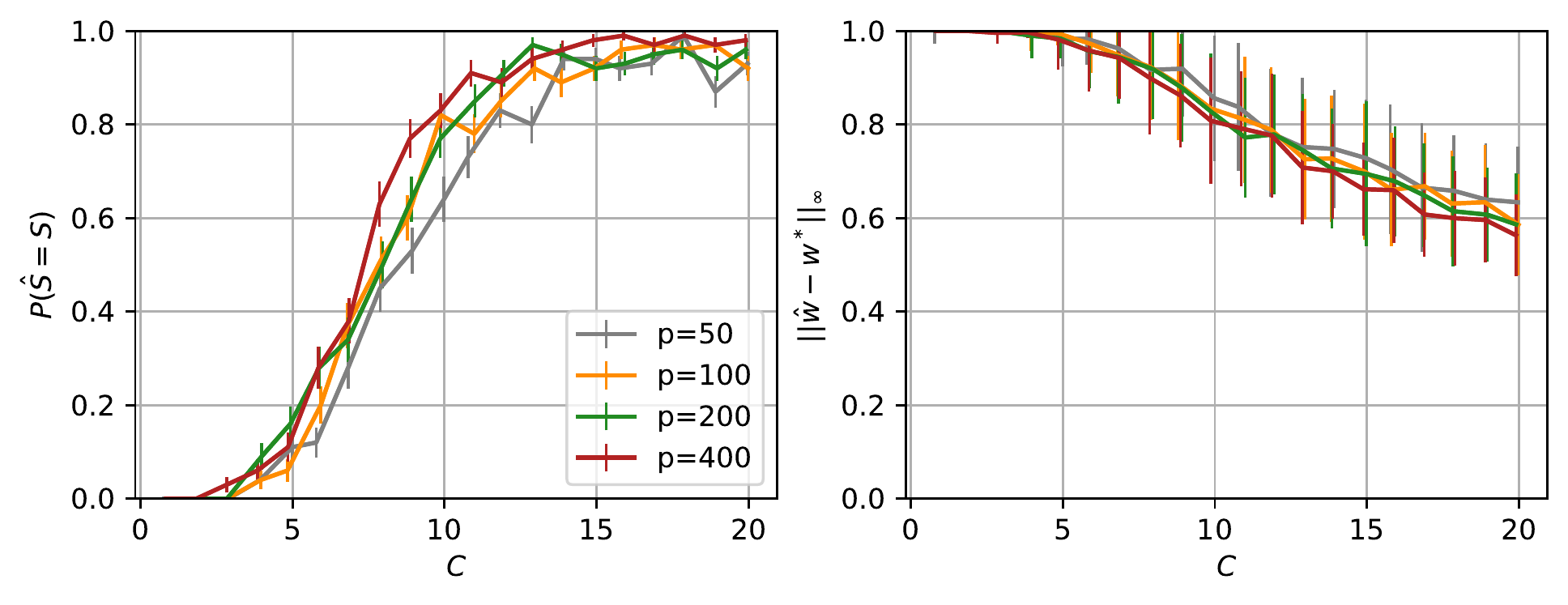}
    \caption{
        Simulations with our meta sparse regression under Gaussian distributions of $\epsilon_{t_i,j}, X_{t_i,j,m}\ \forall i\in[T],j\in[l],m\in S$ and a mixture of sub-Gaussian distributions of $\Delta_{t_i,m}^*\ \forall i\in[T],m\in S$. We use $\lambda=4\sqrt{\frac{\log p }{Tl}}$.
        \textbf{Left:} Probability of exact support recovery for different number of tasks under various settings of number of parameters $p$. The x-axis is set by $C:=\frac{Tl}{k\log(p-k)}$.  \textbf{Right:} The corresponding estimation error of the common parameter $\mathbf{w}$ in $\ell_\infty$ norm. 
    } \label{fig:gaussian_mix_change_p}
\end{figure}

Then we consider the setting of different number of parameters $p$. We choose $p\in\{50,100,200,400\}$ and use $\lambda=4\sqrt{\log p/(Tl)}$ for all the pairs of $(T,l)$. The distribution setting is same as in Figure \ref{fig:gaussian_mix_change_l}. We set $l=5$, and $\mathbf{w}^* = (2,2,2,2,2,0,0,\cdots,0)$. The results are shown in Figure \ref{fig:gaussian_mix_change_p}. The number of tasks $T$ is rescaled to $C$ defined by $\frac{Tl}{k\log(p-k)}$. For different choices of $p$, the curves overlap with each other perfectly (for both $P(\hat{S}=S)$ and $\|\hat{\mathbf{w}} - \mathbf{w}^*\|_\infty$).

{
\subsubsection{Gaussian distribution setting with entries in $X$ being correlated} 

In this section, we consider three different correlation settings in $X$ for our method: 
\begin{enumerate}
    \item $X_{t_i,j}$ are i.i.d. from $N(0,\Sigma_x)$ where $\Sigma_x$ is not a diagonal matrix;
    \item $X_{t_i,j}$ are i.i.d. from $N(0,\Sigma_{x,t_i})$, and $\Sigma_{x,t_i}\sim F(\Sigma)$, i.e., for each task, the covariance matrix of $X$ is different and sampled from a matrix distribution $F(\Sigma)$;
    \item $X_{t_i,j}$ are i.i.d. from $N(\Delta_{t_i}^*,\Sigma_{x,t_i})$, and $\Sigma_{x,t_i} \sim F_{\Delta_{t_i}^*}(\Sigma)$, i.e., for each task, the covariance matrix of $X$ depends on the task specific coefficient $\Delta_{t_i}^*$.
\end{enumerate} 

First, we consider the setting of a nondiagonal $\Sigma_x$ which leads to $\gamma < 1$ in the mutual incoherence condition, where $\gamma = 1-|||\Sigma_{S^c, S}(\Sigma_{S, S})^{-1}|||_{\infty}$. For the simulations we present in the previous sections, the entries in $X_{t_i,j}$ are independent, therefore the covariance matrix of $X_{t_i,j}$ is diagonal and the corresponding $\gamma=1$. Here we consider the case that the entries in $X_{t_i,j}$ are not independent. We choose $p=100,l=5$ and use $\lambda=\sqrt{\log p/(Tl)}$ for all the pairs of $(T,l)$. We set $\mathbf{w}^*$ with five entries equal to 1, and the rest of the entries being 0. The support of $\Delta_{t_i}^*$ is same as the support of $\mathbf{w}^*$. For all $i\in[T],j\in[l],m\in S$, we set $\epsilon_{t_i,j} \sim N(\mu=0,\sigma_\epsilon=0.1)$, $\Delta_{t_i,m}^* \sim N(\mu=0,\sigma_\Delta=0.2)$, $X_{t_i,j} \sim N(\mu=0,\Sigma=\Sigma_x)$, which are mutually independent. The covariance matrix $\Sigma_x=A^T A$ where $A$ is a sum of a randomly generated orthonormal matrix $U_0$ and a matrix $U_1$ with each entry i.i.d. from $\text{Uniform}(-0.05,0.05)$, i.e., $A=U_0+U_1$. After we generate $\Sigma_x$, we calculate the corresponding $\gamma$. We generate $5$ different $\Sigma_x$ with $5$ different $\gamma$. The results are shown in Figure \ref{fig:gaussian_correlated_change_seed}. The number of tasks $T$ is rescaled to $C$ defined by $\frac{Tl}{k\log(p-k)}$.

\begin{figure}[!htbp]
    \centering
    \includegraphics[width=0.75\textwidth]{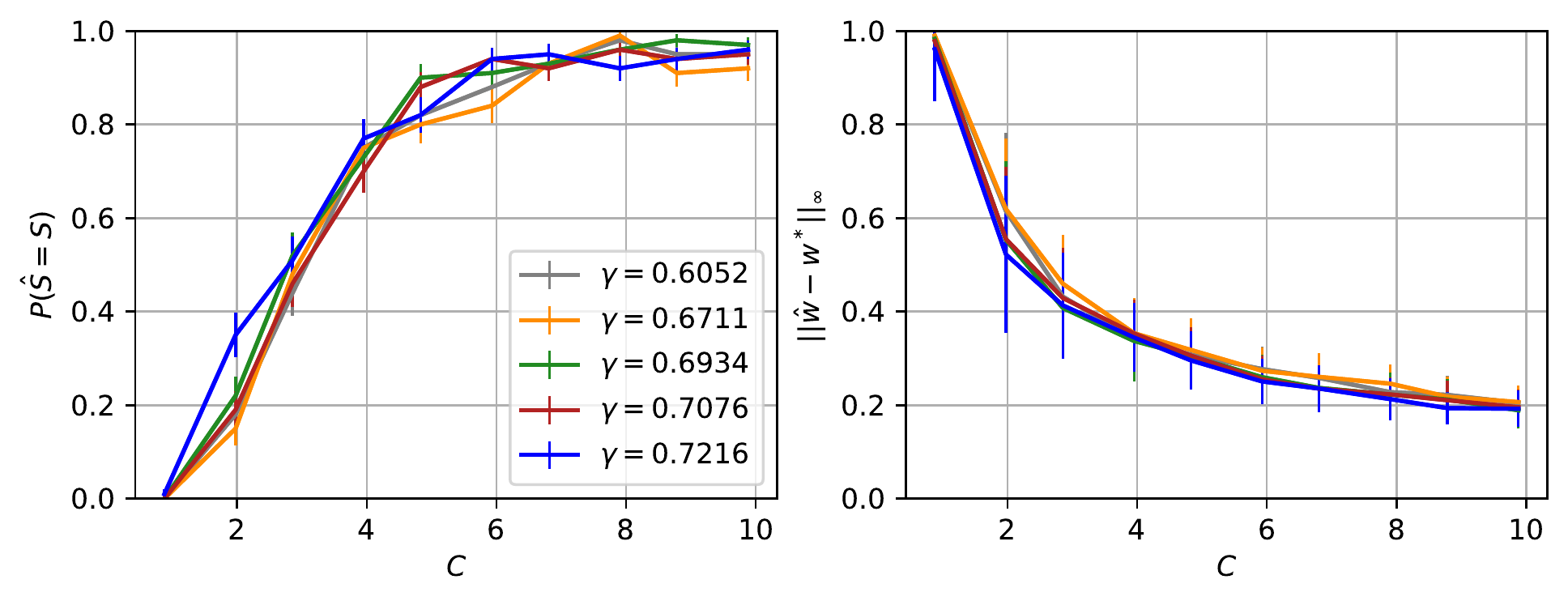}
    \caption{
        Simulations with our meta sparse regression under Gaussian distributions of $\epsilon_{t_i,j}, \Delta_{t_i,m}^*$ and multivariate Gaussian distribution of $X_{t_i,j,m},\ \forall i\in[T],j\in[l],m\in S$. We use $\lambda=\sqrt{\frac{\log p }{Tl}}$.
        \textbf{Left:} Probability of exact support recovery for different number of tasks under various settings of $\gamma$ in the mutual incoherence condition, i.e., $\gamma = 1-|||\Sigma_{S^c, S}(\Sigma_{S, S})^{-1}|||_{\infty}$. The x-axis is set by $C:=\frac{Tl}{k\log(p-k)}$.  \textbf{Right:} The corresponding estimation error of the common parameter $\mathbf{w}$ in $\ell_\infty$ norm. 
    } \label{fig:gaussian_correlated_change_seed}
\end{figure}

Then, we consider the setting of different $\Sigma_{x}$ for each task, i.e., $X_{t_i,j}\sim N(0,\Sigma_{x,t_i}), \Sigma_{x,t_i}\sim F(\Sigma)$. We choose $p=100,l=5$ and use $\lambda=2.5\sqrt{\log p/(Tl)}$ for all the pairs of $(T,l)$. We set $\mathbf{w}^*$ with five entries equal to 1, and the rest of the entries being 0. The support of $\Delta_{t_i}^*$ is same as the support of $\mathbf{w}^*$. For all $i\in[T],j\in[l],m\in S$, we set $\epsilon_{t_i,j} \sim N(\mu=0,\sigma_\epsilon=0.1)$, $\Delta_{t_i,m}^* \sim N(\mu=0,\sigma_\Delta=0.2)$, $X_{t_i,j} \sim N(\mu=0,\Sigma=\Sigma_{x,t_i})$, which are mutually independent. For each task, the covariance matrix $\Sigma_{x,t_i}=A_{t_i}^T A_{t_i}$ where $A_{t_i}$ is a sum of a randomly generated orthonormal matrix $U_{0,t_i}$ and a perturbation matrix $U_{1,t_i}$ with each entry i.i.d. from $\text{Uniform}(-a,a)$, i.e., $A_{t_i}=U_{0,t_i}+U_{1,t_i}, [U_{1,t_i}]_{j,k} \sim\text{Uniform}(-a,a)$. We choose the perturbation range $a$ from $\{0.2,0.1,0.05,0.01\}$. The results are shown in Figure \ref{fig:gaussian_correlated_diff_each_task}. The number of tasks $T$ is rescaled to $C$ defined by $\frac{Tl}{k\log(p-k)}$.

\begin{figure}[!htbp]
    \centering
    \includegraphics[width=0.75\textwidth]{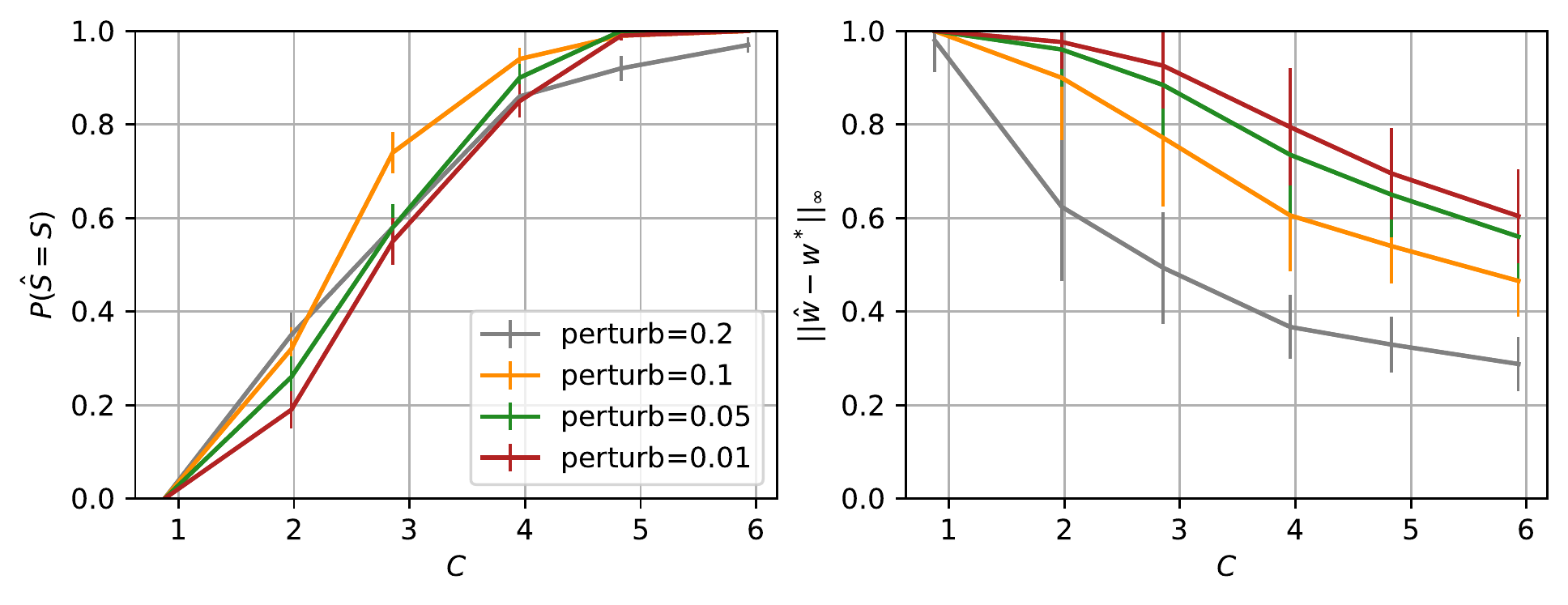}
    \caption{
        Simulations with our meta sparse regression under Gaussian distributions of $\epsilon_{t_i,j}, \Delta_{t_i,m}^*$ and $X_{t_i,j,m}\sim N(0,\Sigma_{x,t_i}),\ \forall i\in[T],j\in[l],m\in S$. We use $\lambda=2.5\sqrt{\frac{\log p }{Tl}}$.
        \textbf{Left:} Probability of exact support recovery for different number of tasks under various settings of $\Sigma_{x,t_i}$ where $\Sigma_{x,t_i}=A_{t_i}^T A_{t_i}, A_{t_i}=U_{0,t_i} + U_{1,t_i}$, $U_{0,t_i}$ is randomly generated orthonormal matrix, $U_{1,t_i}$ is perturbation matrix with each entry i.i.d. from $\text{Uniform}(-a,a)$, i.e., $[U_{1,t_i}]_{j,k} \sim\text{Uniform}(-a,a)$. We choose the perturbation range $a$ from $\{0.2,0.1,0.05,0.01\}$. The x-axis is set by $C:=\frac{Tl}{k\log(p-k)}$.  \textbf{Right:} The corresponding estimation error of the common parameter $\mathbf{w}$ in $\ell_\infty$ norm. 
    } \label{fig:gaussian_correlated_diff_each_task}
\end{figure}

Finally, we consider the setting that for each task, the distribution of $X_{t_i,j}$ depends on the task specific coefficient $\Delta_{t_i}^*$. We choose $p=100,l=5$ and use $\lambda=1.5\sqrt{\log p/(Tl)}$ for all the pairs of $(T,l)$. We set $\mathbf{w}^*$ with five entries equal to 1, and the rest of the entries being 0. The support of $\Delta_{t_i}^*$ is same as the support of $\mathbf{w}^*$. For all $i\in[T],j\in[l],m\in S$, we set $\epsilon_{t_i,j} \sim N(\mu=0,\sigma_\epsilon=0.1)$, $\Delta_{t_i,m}^* \sim N(\mu=0,\sigma_\Delta=0.2)$, which are mutually independent. For each task, $X_{t_i,j} \sim N(\mu=\Delta_{t_i}^*,\Sigma=\Sigma_{x,t_i})$, and the covariance matrix $\Sigma_{x,t_i}=A_{t_i}^T A_{t_i}$ where $A_{t_i}$ is a sum of a randomly generated orthonormal matrix $U_{0,t_i}$ and a perturbation matrix $U_{1,t_i}=a \Delta_{t_i}^* (\Delta_{t_i}^*)^T$, 
i.e., $A_{t_i}=U_{0,t_i}+U_{1,t_i}$. We choose the perturbation range $a$ from $\{0.2,0.1,0.05,0.01\}$. The results are shown in Figure \ref{fig:gaussian_correlation_x_delta_each_task}. The number of tasks $T$ is rescaled to $C$ defined by $\frac{Tl}{k\log(p-k)}$.

\begin{figure}[!htbp]
    \centering
    \includegraphics[width=0.75\textwidth]{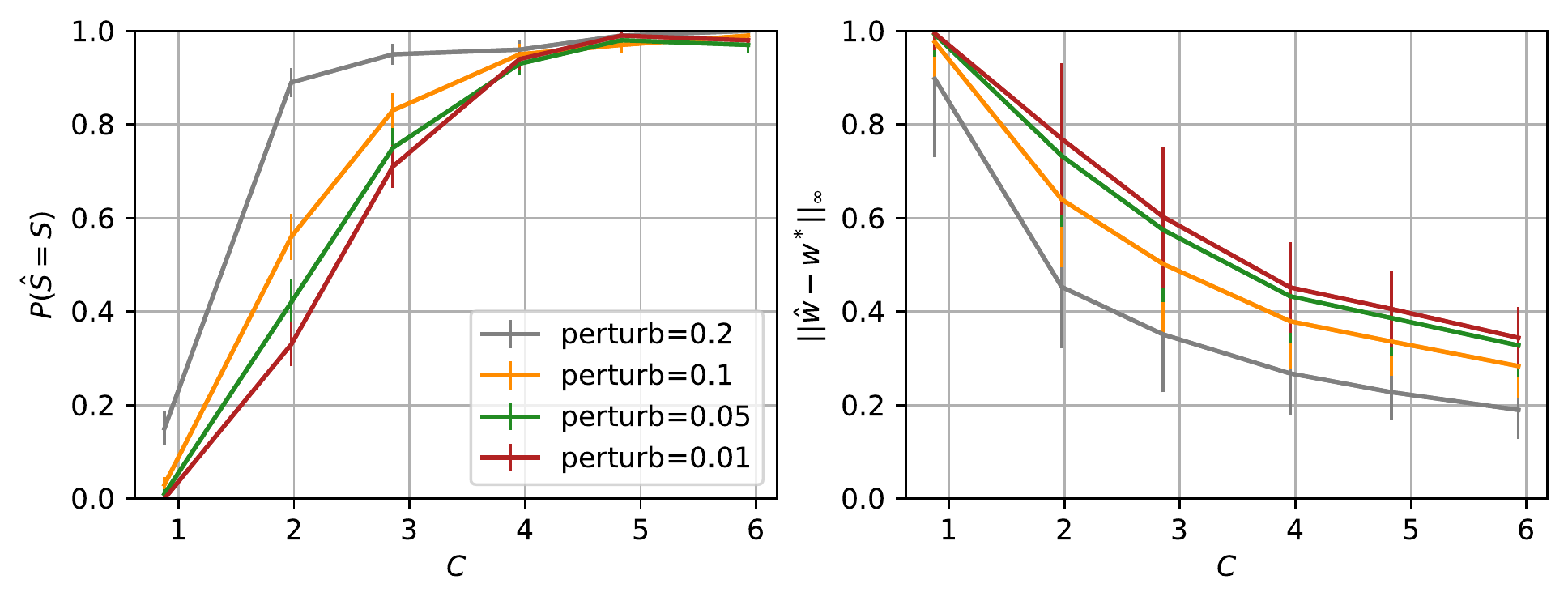}
    \caption{
        Simulations with our meta sparse regression under Gaussian distributions of $\epsilon_{t_i,j}, \Delta_{t_i,m}^*$ and $X_{t_i,j,m}\sim N(\Delta_{t_i}^*,\Sigma_{x,t_i}),\ \forall i\in[T],j\in[l],m\in S$. We use $\lambda=1.5\sqrt{\frac{\log p }{Tl}}$.
        \textbf{Left:} Probability of exact support recovery for different number of tasks under various settings of $\Sigma_{x,t_i}$ where $\Sigma_{x,t_i}=A_{t_i}^T A_{t_i}, A=U_{0,t_i} + U_{1,t_i}$, $U_{0,t_i}$ is randomly generated orthonormal matrix, $U_{1,t_i}=a \Delta_{t_i}^* (\Delta_{t_i}^*)^T$ is perturbation matrix with $a$ from $\{0.2,0.1,0.05,0.01\}$. The x-axis is set by $C:=\frac{Tl}{k\log(p-k)}$.  \textbf{Right:} The corresponding estimation error of the common parameter $\mathbf{w}$ in $\ell_\infty$ norm. 
    } \label{fig:gaussian_correlation_x_delta_each_task}
\end{figure}
}

\subsection{Real-world experiments with a gene expression dataset} 

The single-cell gene expression dataset from \cite{kouno2013temporal} contains expression levels of $45$ transcription factors measured at $8$ distinct time-points. This dataset contains $120$ single cells for each time-point and was used in the experimental validation of \cite{ollier2017regression}. The original objective is to determine the associations among the transcription factors and how they vary over time. We formulate this as a meta-learning problem by setting the first $7$ of the $8$ time-points as the $T$ tasks (for training) and the $8$-th time-point as the novel task (for testing), i.e., $T=7$. Similar to the analysis in \cite{ollier2017regression}, we pick one particular transcription factor, EGR2, as the response variable $y$, and the other $44$ factors as the covariates in $X$, i.e., $p=44$. The true value of the support size $k$ is unknown. We choose $l\in\{5,7,10,15\}$ to model this problem as few-shot learning.

We first randomly permute the $120$ single cells (i.e., samples) while keeping their relative order in all of the $8$ time points (i.e., tasks). Then we find a good choice of hyperparameters: $\lambda$ in our method, 
$\lambda_{1,2}$ for the $\ell_{1,2}$ norm of the method in \cite{obozinski2011support}; $\lambda_1$ and $\lambda_{1,\infty}$ for the $\ell_1$ and $\ell_{1,\infty}$ norms, respectively of the method in \cite{jalali2010dirty}. We use the
tree-structured Parzen estimator approach (TPE) optimizing the criterion of expected improvement (EI) in the Python package \texttt{hyperopt} \cite{bergstra2013making}. 

The search space is $[0,100]$ for all these hyperparameters.
For one choice of the hyperparameters, we choose $l$ samples in each of the $7$ tasks as training samples, and choose the rest $(120-l)$ samples as validation samples. The TPE-EI algorithm evaluates $30$ choices of hyperparameters to minimize the mean square error of the prediction on the validation samples.

\begin{figure}[!htbp]
    \centering
    \includegraphics[width=0.75\textwidth]{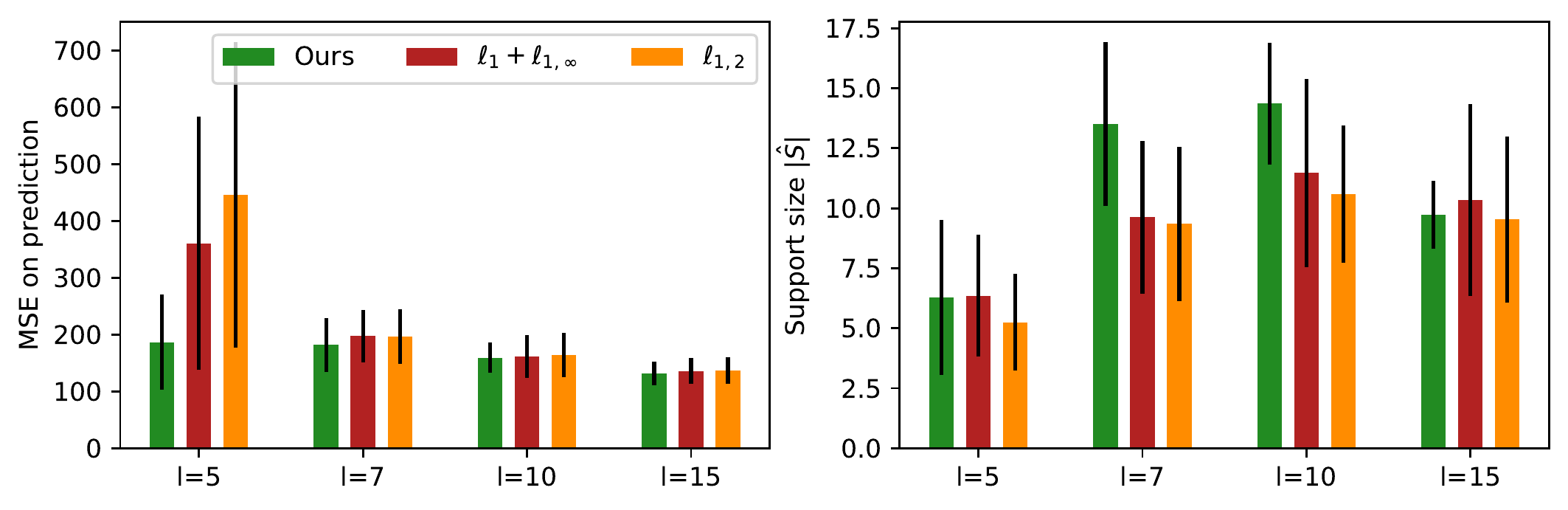}
    \caption{
    Results on the single-cell gene expression dataset from \protect\cite{kouno2013temporal} which was also used in the experimental validation of \protect\cite{ollier2017regression}. Here $T=7, p=44, l\leq 120$. We use our meta learning method and two other multi-task methods to estimate the common support of the $T=7$ tasks and use it to model the data of the novel task. 
    \textbf{Left:}
    The mean square error (MSE) of prediction on the new task.
    \textbf{Right:} 
    The size of the estimated common support $\hat{S}$.
    } \label{fig:real_data}
\end{figure}

After we determine the hyperparameters from all the three methods (ours, $\ell_{1,2}$, and $\ell_{1}+\ell_{1,\infty}$), we choose $l$ samples in each of the $7$ tasks to train models by these methods to estimate $S$ (for multi-task methods, $\hat{S}:=\bigcup_{i=1}^T \hat{S}_i$.) The mean and standard deviation of the size of the estimated support are shown in the right panel of Figure \ref{fig:real_data}.

When the estimated common supports are obtained, we can use LASSO constrained on the common support to solve for the new task, i.e., the $8$-th time point. We determine the choice of hyperparameters using \texttt{hyperopt} in the same way shown above. Then we use LASSO with $\lambda$ being set to those hyperparameters to estimate the support of the new task. Since the weight estimation of $(\mathbf{w}^* + \Delta_{t_{T+1}}^*)$ by LASSO is not very accurate when the sample size $l$ is small, we use linear regression to estimate $(\mathbf{w}^* + \Delta_{t_{T+1}}^*)$ again with the support recovered by LASSO. The performance is measured by the mean square error (MSE) of prediction on the rest $(120-l)$ samples. For one estimated common support, we take $6$ random choices of the training $l$ samples in the new task and calculate the mean of the the prediction error. The mean and standard deviation of MSE are shown in the left panel of Figure \ref{fig:real_data}.

All the mean and standard deviation results (shown as error bars) in Figure \ref{fig:real_data} are obtained from $100$ repetitions of the experiment setting above. From Figure \ref{fig:real_data} we can see that our method has lower MSE when $l$ is small. Since $T$ is not large and does not grow, the multi-task methods also perform well when $l$ is large enough. We also show that the size of the estimated common support by our methods is not significantly larger than the ones by the other two multi-task methods, which suggests that our method produces a more accurate estimation of the common support set.

\end{document}